\documentclass[journal, letterpaper]{IEEEtran}
\usepackage{graphicx}
\usepackage{url}        
\usepackage{amsmath}    
\usepackage{textgreek}	
\usepackage{listings}
\usepackage{csvsimple}
\usepackage{longtable}
\newcommand{\argmin}{\mathop{\rm arg~min}\limits}

\usepackage{subcaption}
\usepackage{pgfplotstable}
\usepackage{tikz}
\usepackage[flushleft]{threeparttable}
\usepackage{float}
\usepackage{multirow}
\begin{document}

\title{Efficient Stereo Matching on Embedded GPUs with Zero-Means Cross Correlation}
\author{Qiong Chang, Aolong Zha, Weimin Wang, Xin Liu, Masaki Onishi, Lei Lei, Meng Joo Er, Tsutomu Maruyama}
\maketitle

\begin{abstract}
Mobile stereo-matching systems have become an important part of many
applications, such as automated-driving vehicles and autonomous robots.
Accurate stereo-matching methods usually lead to high computational complexity;
however, mobile platforms have only limited hardware resources to keep their
power consumption low; this makes it difficult to maintain both an acceptable
processing speed and accuracy on mobile platforms.
To resolve this trade-off, we herein propose a novel acceleration approach for
the well-known zero-means normalized cross correlation (ZNCC) matching cost
calculation algorithm on a Jetson Tx2 embedded GPU. In our method for
accelerating ZNCC, target images are scanned in a zigzag fashion to efficiently
reuse one pixel's computation for its neighboring pixels;
this reduces the amount of data transmission and increases the utilization of
on-chip registers, thus increasing the processing speed.
As a result, our method is 2X faster than the traditional image scanning method,
and 26\% faster than the latest NCC method.
By combining this technique with the domain transformation (DT) algorithm, our
system show real-time processing speed of 32 fps, on a Jetson Tx2 GPU for
1,280x384 pixel images with a maximum disparity of 128.
Additionally, the evaluation results on the KITTI 2015 benchmark show that our
combined system is more accurate than the same algorithm combined with census by
7.26\%, while maintaining almost the same processing speed. Source Code:
\textit{https://github.com/changqiong/Z2ZNCC.git}.
\end{abstract}

\section{INTRODUCTION}
\label{sec:1}
Stereo matching is a key algorithm for depth detection in
computer vision, but its usability is still limited because attaining high
accuracy requires a very high computational complexity.
By achieving both high accuracy and processing speed on mobile platforms, it can
be used in many applications, including auto-driving, autonomous robots, and so
on.
Thus far, many researchers have focused upon accelerating stereo matching on
mobile platforms.
Most of them focus on accelerating the two most computationally intensive
stages: cost calculation and cost aggregation.
During cost calculation, each pixel in the reference image is first matched with
several pixels in the target image one by one. Next, the similarity between any
two pixels is quantified by a numerical value (cost) calculated by a matching
method, such as the sum of absolute differences (SAD), census, or convolution
neural network (CNN).
Then, to further improve the accuracy of the matching system, the cost of each
pixel within a certain region is expected to be aggregated together to represent
the similarity between any two regions; this is called matching-cost
aggregation.
Many methods can be used to determine the range of the matching regions, such as
semi-global matching (SGM) or domain transformation (DT).
Various combinations of the above two stages not only result in different
matching accuracies, but also different computational complexities; this leads
to differing processing speeds.
Many researches implement their stereo-matching systems on FPGAs,
respectively. Wang~\cite{r21} first combines a simplified SGM with the simple
absolute differences and census matching algorithms, processing 1,024x768 pixel
images with 96 disparities at 67 fps.
Mohammad~\cite{r1}, Zhang~\cite{r23} and Kuo~\cite{r24} use the census algorithm
to calculate their matching costs. Mohammad~\cite{r1} combines census with a
cross-aggregation method to achieve a good error rate of less than 9.22\% and a
high processing speed of faster than 200 fps on the KITTI 2015
benchmark~\cite{r6}. Zhang~\cite{r23} uses a box filter to aggregate matching
costs and achieves a high processing speed of 60 fps for 1080p
images. Kuo~\cite{r24} uses a two-pass aggregation method and achieves the same
processing speed as ~\cite{r23}.
Oscar~\cite{r2} uses SAD to calculate the matching cost and combines it with
SGM. Due to SGM's high accuracy for even the simple SAD matching algorithm, it
can still achieve a lower error rate of 8.7\%, except that its speed is reduced
to 50 fps.
Additionally, Zhang~\cite{r22} develops a special ASIC to accelerate the
implementation of SGM and achieves a processing speed of 30fps for 1080p images.
Here, due to the limitations of floating-point decimal calculation, both the
FPGA-based and the dedicated ASIC-based systems usually use methods such as SAD
and census to obtain integer cost values. Although this is conducive to
implementation on them, it also limits improvement in the matching accuracy.
Furthermore, hardware-based systems typically need long development cycles and
are also difficult to maintain.
The recent advent of embedded GPUs has allowed the development of many
systems~\cite{r25}~\cite{r26}. Compared to FPGAs, embedded GPU-based systems
have short development cycles~\cite{r27}. In addition, they are easy to maintain
and port on other platforms. Wang \cite{r3}, Smolyanskiy~\cite{r4} and
Tonioni~\cite{r5} implement their systems on a Jetson Tx2 embedded GPU using the
CNN;
according to the evaluation results of KITTI 2015~\cite{r6} benchmark, their
accuracies are high, with error rates between 3.2\% and 6.2\%.
However, due to the significant calculations of CNN-based methods, their
processing speeds are only a few fps, far below the requirements for practical
applications.
Daniel~\cite{r7} constructs a fast stereo-matching system on a Jetson Tx2 GPU. It
also combines census algorithm with SGM to achieve an error rate of 8.66\% and a
processing speed of 29 fps on KITTI 2015 benchmark. It is currently the best
system for balancing the accuracy and processing speed on mobile GPUs;
however, due to the census-matching method, this system is still not accurate
enough, even if implemented on GPUs that are good at floating-point decimal
calculations.

According to~\cite{r8}, the matching accuracy by normalized cross correlation
(NCC) is better than that by census because it has a higher ability to withstand
changes in gain and bias.
Furthermore, zero-means NCC (ZNCC)--an improved version of NCC--provides strong
robustness because it also tolerates uniform brightness variations
\cite{r9}\cite{r10}.
However, ZNCC has not been widely used on the mobile systems with
limited hardware resources because of its higher computational
complexity.

In this paper, we accelerate ZNCC on a Jetson Tx2 embedded GPU and make it
possible to achieve a comparable processing speed to that of census with a
higher matching accuracy. The main contributions of this paper are as follows:
\begin{itemize}
\item We introduce a new calculation method, zigzag scanning based zero-means
  normalized cross correlation (Z$^2$-ZNCC) to reuse the computational results
  of a pixel for the calculations of its neighbors. This makes it possible to
  reduce data transfer between the global memory of the GPU and increase the
  processing speed. 
\item We propose a strategy to make efficient use of registers during zigzag
  scanning to achieve higher parallelism of GPU threads driven by GPU cores.
\item We design GPU-implementation algorithms for two parallel summation methods
  used in our Z$^2$-ZNCC and comprehensively compare their performance.
\item We create FastDT, an upgraded version of the GPU-based domain
  transformation method presented in~\cite{r13} by removing the cost-value
  shifting step and increasing the flag code. Then, we combine it with
  Z$^2$-ZNCC to construct a real-time stereo-matching system on an embedded GPU.
\end{itemize}
The experimental results demonstrate that our method is 2X faster than the
traditional image-scanning method and 26\% faster than the latest NCC method
\cite{r12}. Furthermore, our system achieves a processing speed of 32 fps and an
error rate of 9.26\% for 1,242x375 pixel images when the maximum disparity is
128 on a KITTI 2015 dataset. It is one of the few embedded GPU-based real-time
systems, with an accuracy much higher than others.

This paper extends our previous work (short paper)~\cite{r20} from the following
aspects.
\begin{itemize}
\item We design two parallel summation methods which maximize the processing
  speed of Z$^2$-ZNCC depending on the template size.
\item We introduce an efficient two-step implementation technique for domain
  transformation, which not only maintains a high processing speed, but also a
  high accuracy of Z$^2$-ZNCC.
\item We conduct comprehensive experiments to examine the impact of various
  conditions on the processing speed of Z$^2$-ZNCC.
\end{itemize}

The rest of the paper is organized as follows. Section~\ref{sec:2} introduces
the related works. Section~\ref{sec:3} reviews ZNCC and DT calculation
methods. Section~\ref{sec:4} discusses the GPU implementation of Z$^2$-ZNCC and
FastDT. Section~\ref{sec:5} shows the evaluation results. Finally,
Section~\ref{sec:6} presents the conclusions and our future work.

\section{Related works}
\label{sec:2}
Recently, many researchers have focused upon accelerating the performance of
NCC-based methods and applying them into stereo-matching systems.

Lin~\cite{r14} proposes an optimization method for ZNCC calculation on a
general platform. This method divides the standard equation into four
independent parts and calculates the correlation coefficient efficiently using
sliding windows. Hence, the computational complexity of ZNCC could be reduced
from the original order.
The computation time becomes constant for the window size; however, a large
memory is required to store the calculation results for reuse.  Although this
method works nearly 10X faster than traditional ones, it is not applicable to
embedded GPUs because of their limited memory space.
Rui~\cite{r15} implements a fast ZNCC-based stereo-matching system on GTX 970M
GPU. This work focuses on the use of integral images to calculate the mean and
standard deviation efficiently. Rui's system runs approximately 9X faster than a
single-threading CPU implementation and about 2X faster than eight-threading;
however, according to our evaluation, this approach
is inefficient for a small-size window (less than 9x9) because obtaining an
integral image itself also requires calculation costs.
These costs are mainly caused by data transfer of the integration results, which
cannot be ignored for an embedded GPU with a high memory latency.
Han~\cite{r12} implements an NCC-based stereo-matching system on a Jetson Tx2
GPU. This method divides the equation into three parts, each with an identical
control flow but different data locations. All intermediate results are stored
on shared memory evenly so as to accelerate the calculation through
reuse. However, as mentioned in Section \ref{sec:1}, the heavy use of shared
memory reduces the parallelism of GPU blocks.

\section{Algorithms and Optimizations}
\label{sec:3}
\subsection{Zero-means Normalized Cross Correlation (ZNCC)}
\label{sec:3_1}
ZNCC is used to calculate matching costs between a reference pixel $I_R(x,y)$ in
the reference image and a series of target pixels $I_T(x-d,y)$ in the target
image. $d$ is called {\itshape disparity}, and its range is $[0,D)$, where $D$
  is a constant called maximum disparity. The function of ZNCC is given as
  follows:\vspace*{-0.1cm}
  
{\small
\begin{equation}
C(x,y,d) = \frac{\displaystyle
  \sum_{(x,y)\in{W}}{\Delta{I_{R}}(x,y)\cdot{\Delta{I_T}(x-d,y)}}}{\sigma_{R}(x,y)\cdot{\sigma_{T}(x-d,y)}},
\label{equ:ZNCC}
\end{equation}
\vspace*{-0.1cm}}
where
\vspace*{-0.2cm}
{\small
\begin{equation}
\begin{split}
\sigma_{R}(x,y) &= \sqrt{\displaystyle \sum_{(x,y)\in{W}}{\Delta{I_R(x,y)^2}}},\nonumber \\
\sigma_{T}(x-d,y) &= \sqrt{\displaystyle \sum_{(x,y)\in{W}}{\Delta{I_T(x-d,y)^2}}},\nonumber \\
\end{split}
\end{equation}
\vspace*{-0.2cm}}
and
\vspace*{-0.2cm}
{\small
\begin{equation}
\begin{split}
\Delta{I_{R}(x,y)} &= I_{R}(x,y)-\overline{I_{R}(x,y)}, \nonumber \\
\Delta{I_{T}(x-d,y)} &= I_{T}(x-d,y)-\overline{I_{T}(x-d,y)}. \nonumber \\
\end{split}
\end{equation}
}
Here, $\overline{I_{R}(x,y)}$ and $\overline{I_{T}(x-d,y)}$ are the averages of
the pixel values in the matching windows $W$ surrounding $I_{R}(x,y)$ and
$I_{T}(x-d,y)$, respectively.
$C(x,y,d)$ in \eqref{equ:ZNCC} is the correlation coefficient (i.e., the
matching cost) between $I_R(x,y)$ and $I_T(x-d,y)$; its range is $[0,1]$ (the
closer to one, the more similar the two windows).
$\sigma_R(x,y)$ and $\sigma_T(x-d,y)$ are the standard deviations of the pixel
values in the two windows and are used to normalize the correlation coefficient
between them.
Each reference pixel $I_R(x,y)$ needs to be matched with $D$ target
pixels $I_T(x-d,y)$; 
here, $\Delta{I_R(x,y)}$ and $\Delta{I_T(x-d,y)}$ can be calculated in advance
because the calculations of these terms are closed in each image.
As such, the total number of calculations can be reduced. However, when the size
of the matching window $W$ is $l^2$ (where $l=2r+1$ represents the side length
of window $W$) $l^2$-times the memory space is needed for each image because
each window has $l^2$ differences.
To further reduce the ZNCC's computation complexity, \eqref{equ:ZNCC} can be
rewritten as follows:\vspace*{-0.2cm}

{\small
\begin{equation}
\begin{split}
C(x,y,d)=\frac{\displaystyle
  \sum_{(x,y)\in{W}}{\Pi_{RT}(x,y,d)}-l^2\cdot{\overline{\Pi_{RT}(x,y,d)}}}{\sigma_{R}(x,y)\cdot{\sigma_{T}(x-d,y)}},
\end{split}
\label{equ:OP_ZNCC}
\end{equation}
}
where
{\small
\begin{equation}
\begin{split}
\Pi_{RT}(x,y,d) &= I_R(x,y)\cdot{I_T(x-d,y)}, \nonumber \\ 
\overline{\Pi_{RT}(x,y,d)} &= \overline{I_R(x,y)}\cdot{\overline{I_T(x-d,y)}}, \nonumber \\
\end{split}
\end{equation}
}
and
{\small
\begin{equation}
\begin{split}
\sigma_{R}(x,y) &= \sqrt{\displaystyle
  \sum_{(x,y)\in{W}}{I_R(x,y)^2}-l^2\cdot{\overline{I_R(x,y)}^2}},\nonumber
\\ \sigma_{T}(x-d,y) &= \sqrt{\displaystyle
  \sum_{(x,y)\in{W}}{I_T(x-d,y)^2}-l^2\cdot{\overline{I_T(x-d,y)}^2}}.\nonumber \\
\end{split}
\end{equation}
}
In this calculation, $C(x,y,d)$, $\sigma_{R}(x,y)$, and $\sigma_{T}(x-d,y)$ are
calculated from four values; $\overline{I_R(x,y)}^2$, $\sum{I_R(x,y)^2}$,
$\overline{I_T(x-d,y)}^2$, and $\sum{I_T(x-d,y)^2}$, rather than from
$\Delta{I_R(x,y)}$, and $\Delta{I_T(x-d,y)}$ as shown in \eqref{equ:ZNCC}.
These four values are only related to their respective images and can all be
calculated in advance. Thus, both $\sigma_{R}(x,y)$ and $\sigma_{T}(x-d,y)$ can
be calculated efficiently without calculating $\Delta{I_R(x,y)}$ and
$\Delta{I_T(x-d,y)}$ for each pixel $l^2$ times. This transformation not only
helps to reduce the total calculation amount, but also reduce the required
memory space.
Of course, ZNCC-based matching is performed in a fixed size window, which is not
accurate for irregular patterns in reality; therefore, the matching cost
$C(x,y,d)$ is usually combined with various aggregation methods to improve the
matching accuracy.
\subsection{Domain Transformation (DT)}
\label{sec:3_3}
In cost aggregation step, the matching costs of all similar
  pixels in the same area (e.g., the area within the pink-dashed line in
  Fig.\ref{fig:DT} (a)) are added together.
{\itshape Domain Transformation (DT)}~\cite{r16} is an effective algorithms for
use at this stage.
Unlike other algorithms, DT avoids over-fitting of cost propagation by using the
gradients of adjacent pixels to weight their costs in different directions,
rather than judging the boundary of each area in advance.
The advantage of DT is that there is no need to segment each area for cost
aggregation separately, and it is suitable for parallel processing to increase
the aggregation speed.
In DT, the matching cost of each pixel is aggregated from four different
directions, while propagating its cost to four neighboring pixels is done
according to the following equations:
\begin{figure*}
 \centering \includegraphics[width=7in]{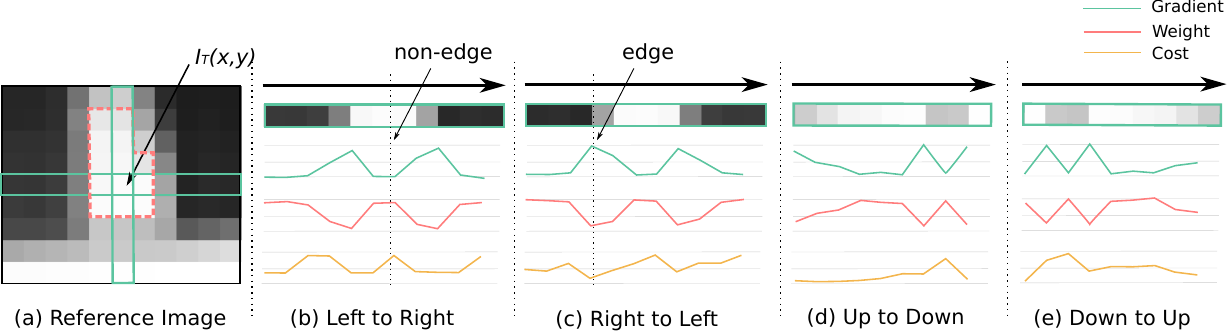}
    \caption{Domain transformation}
    \label{fig:DT}
    \vspace*{-0.4cm}
\end{figure*}
\begin{align}
C_L(x,y,d) &= C(x,y,d) \small{+} C_L(x\small{-}1,y,d)\cdot{W_L(x,y)}\label{equ:DT1},\\
C_R(x,y,d) &= C_L(x,y,d) \small{+} C_R(x\small{+}1,y,d)\cdot{W_R(x,y)}\label{equ:DT2},\\
C_U(x,y,d) &= C_R(x,y,d) \small{+} C_U(x,y\small{-}1,d)\cdot{W_U(x,y)}\label{equ:DT3},\\
C_D(x,y,d) &= C_U(x,y,d) \small{+} C_D(x,y\small{+}1,d)\cdot{W_D(x,y)}\label{equ:DT4}.
\end{align}
Here, $C_L$, $C_R$, $C_U$, and $C_D$ represent the aggregated costs for each
pixel from four different directions (left, right, up, and down).
In this aggregation, for example, the boundary condition for (\ref{equ:DT1}) is
given by $C_L(0,y,d) = C(0,y,d)$.
$W_L$, $W_R$, $W_U$ and $W_D$ represent the corresponding weights calculated by
the gradient between adjacent pixel values according to the following equations:
\begin{equation}
\label{equ:k1}
W_L(x,y) = a^{1+\frac{\sigma_s}{\sigma_r}\cdot{|I_T(x,y)-I_T(x-1,y)|}},
\end{equation}
\begin{equation}
\label{equ:k2}
W_R(x,y) = a^{1+\frac{\sigma_s}{\sigma_r}\cdot{|I_T(x,y)-I_T(x+1,y)|}},
\end{equation}
\begin{equation}
\label{equ:k3}
W_U(x,y) = a^{1+\frac{\sigma_s}{\sigma_r}\cdot{|I_T(x,y)-I_T(x,y-1)|}},
\end{equation}
and
\begin{equation}
\label{equ:k4}
W_D(x,y) = a^{1+\frac{\sigma_s}{\sigma_r}\cdot{|I_T(x,y)-I_T(x,y+1)|}},
\end{equation}
where
\begin{equation}
\label{equ:a}
a = \exp{(-\frac{1}{\sigma_s})}.
\end{equation}
In \eqref{equ:k1} to \eqref{equ:a}, $\sigma_s$ is a spatial parameter and
$\sigma_r$ is an intensity range parameter. Both are used to adjust the weight
caused by gradient changes in space and intensity.
To simplify the calculation, the weight equations can be further simplified
below (here, we only take \eqref{equ:k1} as an example):
\begin{align}
\ln{W_L} &=
\ln{(a^{1+\frac{\sigma_s}{\sigma_r}\cdot{|I_T(x,y)-I_T(x-1,y)|}})}\nonumber
\\ &=(1+\frac{\sigma_s}{\sigma_r}\cdot{|I_T(x,y)-I_T(x-1,y)|})\cdot{\ln{a}}
\nonumber\\ &=(1+\frac{\sigma_s}{\sigma_r}\cdot{|I_T(x,y)-I_T(x-1,y)|})\cdot{(-\frac{1}{\sigma_s})}\nonumber\\ &=-\frac{1}{\sigma_s}-\frac{|I_T(x,y)-I_T(x-1,y)|}{\sigma_r},
\nonumber\\
\end{align}
then
\begin{align}
  \label{equ:weight}
W_L&=\exp(-\frac{1}{\sigma_s}-\frac{|I_T(x,y)-I_T(x-1,y)|}{\sigma_r})\nonumber\\ &=K\cdot{\exp(-\frac{|I_T(x,y)-I_T(x-1,y)|}{\sigma_r})},\nonumber\\
\end{align}
where
\begin{align}
K = \exp(-\frac{1}{\sigma_s}).
\end{align}
$K$ is a constant coefficient used to ease the calculation of $W_L$.
Figure.\ref{fig:DT} shows an example of cost aggregation for pixel
$I_T(x,y)$. Figure.\ref{fig:DT} (a) shows part of the reference image centered on
$I_T(x,y)$.
Figures.\ref{fig:DT} (b) to \ref{fig:DT} (e) show the cost propagation process from
different directions.
Three curves with different colors represent the changes in the gradient,
weights, and propagated costs, respectively.
According to \eqref{equ:weight}, the weight is calculated by the gradient and
then used to weight the propagated cost value.
As shown in Figs.\ref{fig:DT} (b) to \ref{fig:DT} (e), the weight changes in the
opposite direction to the change in gradient value, thereby ensuring that
cost propagation can be performed normally among non-edge pixels
(Fig.\ref{fig:DT} (b)) and can also be interrupted at edge pixels
(Fig.\ref{fig:DT} (c)).
When the propagation from down to up is completed (i.e., when the final
aggregation result $C_D(x,y,d)$ is obtained), $C(x,y,d)$ is replaced by
$C_D(x,y,d)$, and used in the following stages.

\subsection{Winner-Take-ALL (WTA)}
After calculating matching cost for maximum disparity $D$ times, the target
pixel $I_T(x-d,y)$ that is most similar to reference pixel $I_R(x,y)$ is
determined as:
\begin{equation}
  D_{map}(x,y) = \argmin_{d}{(1-C(x,y,d))}.
  \label{equ:WTA1}
\end{equation}
As shown in this equation, the value of $d$ that minimizes $(1-C(x,y,d))$ is
chosen as the disparity of the reference pixel $I_R(x,y)$.

\section{Implementation}
\label{sec:4}
Implementing ZNCC and DT on an embedded GPU is a key challenge for realizing a
fast and accurate mobile stereo-vision system. In this section, we first
introduce the architecture of Jetson Tx2 GPU; then, we describe the acceleration
approaches of ZNCC and DT, respectively.

\subsection{GPU Architecture and CUDA Programming Model}
\label{sec:4_1}
The Jetson Tx2 has 2 streaming multi-processors (SMs); each SM runs in parallel
using 128 cores (256 cores in total) and has two types of on-chip memory:
register memory and shared memory. Their sizes are limited, but their access
latencies are very low.
This GPU also has a global memory (off-chip), which is usually used to hold all
data for processing. Due to the high latency of access to the off-chip memory,
the most important point for achieving high performance on the GPU is to
minimize the amount of data transfer between on-chip and off-chip memory.
In our implementation, we use the GPGPU programming model CUDA~\cite{r17}.
CUDA abstractly defines the GPU core, SM, and GPU itself as thread, block, and
grid, respectively. A grid is composed of blocks and a block is composed of
threads. Every 32 threads execute the same instruction, which is called a
{\itshape warp}.  The warps are scheduled serially by the SMs.
Users can define the number of the abstract resources according to their
requirements, which may exceed the physical GPU resources; then, the CUDA driver
schedules abstract resources to work upon physical resources.
Since the total number of registers and shared-memory space are fixed, the
amount of these resources allocated to each thread and block determines how many
of them can be active. The more allocated, the fewer threads and blocks can be
activated, which resulting in reduced performance.
By storing intermediate calculation results and reusing them afterwards, the
total amount of calculation can be reduced; however, more hardware resources are
needed to store these results, which limits the number of active threads.
On the other hand, by recalculating them each time, the required amount
of hardware resources can be reduced, and more threads can be active.
Balancing the hardware-resource usage and total amount of calculation is a key
point for achieving high performance, especially on embedded GPUs with limited
hardware resources.

\subsection{Implementation of ZNCC}
\label{sec:4_2}
Our ZNCC-acceleration approach includes two steps: (1) calculation of the means
and the sums of squares in matching windows, and (2) calculation of the
correlation coefficients of each pixel by zigzag scanning.
\subsubsection{Summation}
\label{sec:4_2_1}
For the ZNCC-acceleration approach, the means and sums of squares in each
matching window are calculated in advance (Section~\ref{sec:3_1}). Taking the
example of the pixel values in the reference image, we describe the two methods
as follows (for simplicity, we only describe the sum in the reference image):
\begin{figure}
 \centering \includegraphics[width=2.7in]{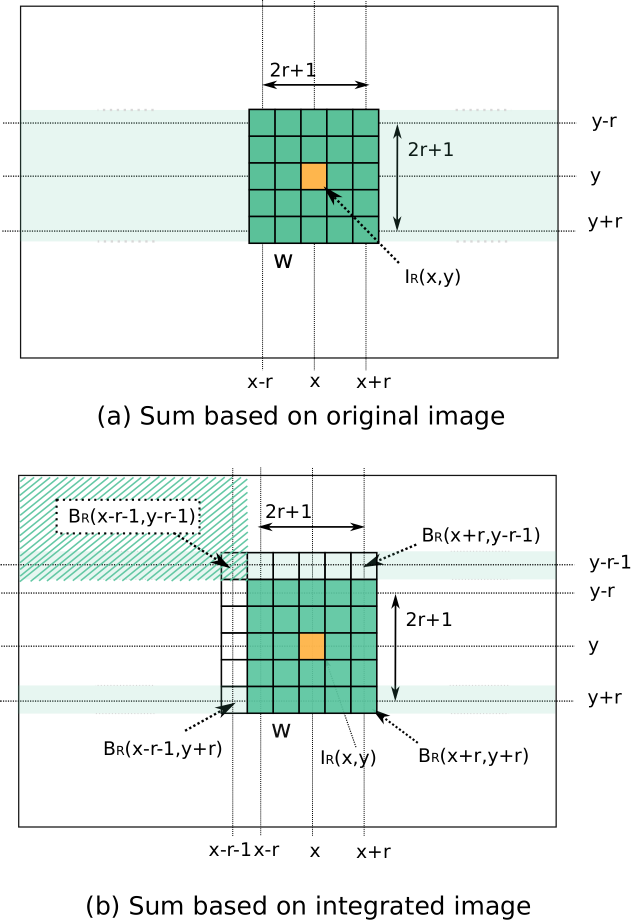}
    \caption{Summation methods}
    \label{fig:Sum}
    \vspace*{-0.4cm}
\end{figure}
{\itshape Method 1:} As shown in Fig.\ref{fig:Sum} (a), $S_{R}(x,y)$, the sum of
the pixel values in each matching window $W$ is calculated as
follows:\vspace*{-0.4cm}

{\small
\begin{equation}
\begin{split}  
S_{R}(x,y) &= \displaystyle \sum_{(x,y)\in W}I_R(x,y). \\
\end{split}
\label{equ:sum1}
\end{equation}
}
Here, $(2r+1)^2$ pixel values are simply added around the center pixel
$I_R(x,y)$.

The summation of each window is performed independently by different CUDA
threads.
Since the two adjacent windows for two adjacent pixels share $2r\times{(2r+1)}$
pixels, the data from $(2r+1)$ rows or columns are excepted to be cached in the
same shared memory.
The allocated memory space grows as the window size increases.
The number of columns and rows processed by each CUDA block at the same time
depends upon the size of the shared memory allocated;
for smaller-sized windows, less hardware resources are required for each thread,
and higher parallelism can be expected; however, as the window size increases,
more hardware resources are required, and fewer threads can be active.
Therefore, this method is not suitable for the summation of large windows.
{\itshape Method 2:} This method performs the summation using the integral image
$B_R(x,y)$:\vspace*{-0.2cm}

{\small
\begin{equation}
\displaystyle B_R(x,y)=\sum_{u=0}^{x}\sum_{v=0}^{y}I_R(u,v).
\end{equation}
}
As shown in Fig.\ref{fig:Sum} (b), $S_R(x,y)$ is calculated using four points in
the integral image, regardless of the window size:\vspace*{-0.2cm }

{\small
\begin{equation}
\begin{split}
  S_R(x,y)&{\small=}
  B_R(x+r,y+r)+B_R(x{\small-}r{\small-}1,y{\small-}r{\small-}1)\\ &{\small-}B_R(x+r,y{\small-}r{\small-}1){\small-}B_R(x{\small-}r{\small-}1,y+r).
\end{split}
\end{equation}
}
In this calculation method, To calculate the sum for the pixels on row $y$, only
two rows ($y-r-1$ and $y+r$) of the integral image are needed.
Thus, this method is suitable for the summation of large windows.

Obtaining an integral image in parallel requires two steps:
\begin{enumerate}
\item integrate each row of $I_R$, defined as $B_H$,
\item integrate each column of $B_H$, defined as $B_R$.
\end{enumerate}

\begin{figure}
 \centering \includegraphics[width=3in]{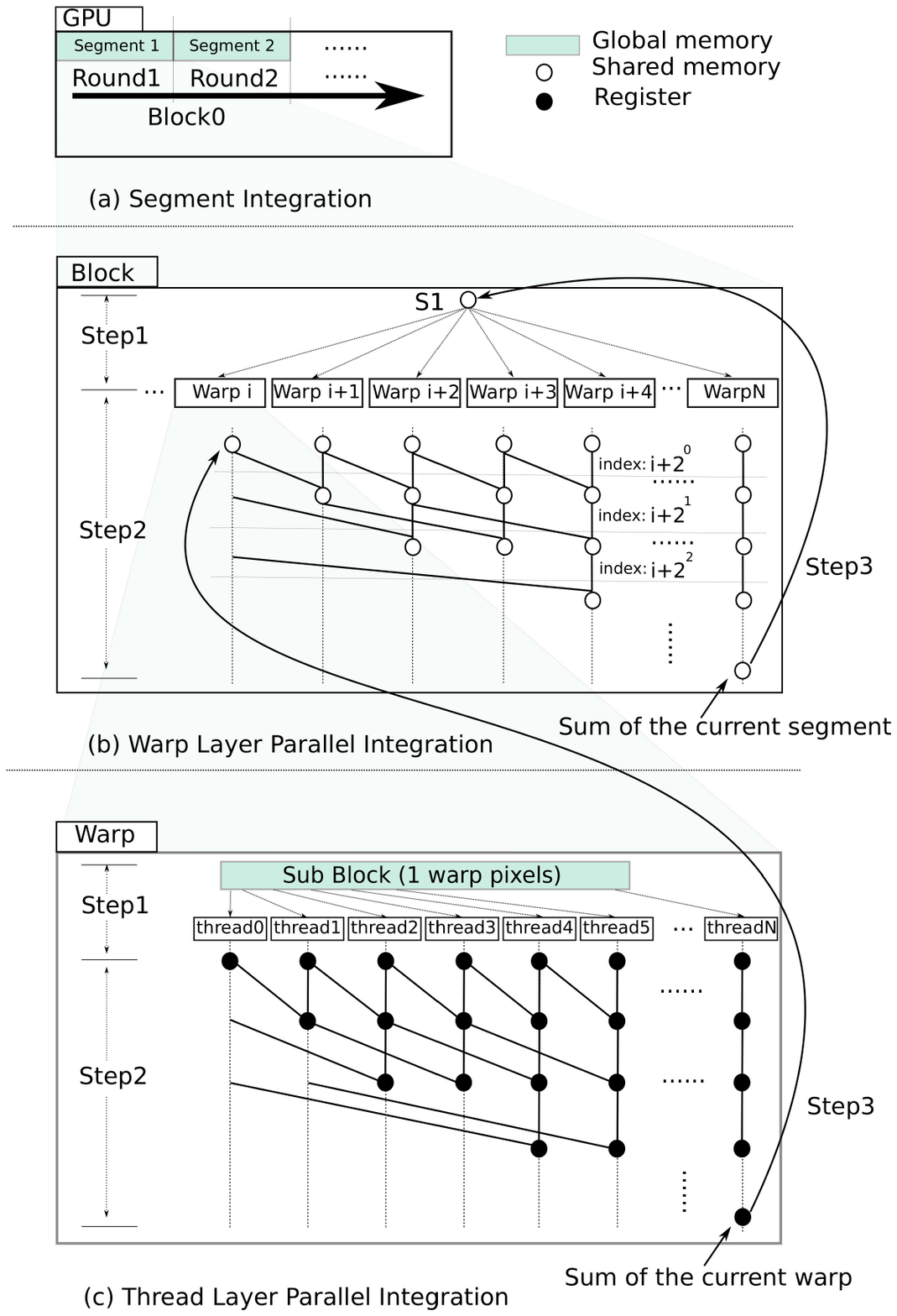}
    \caption{Integration along the $x$-axis}
    \label{fig:Integration}
    \vspace*{-0.5cm}
\end{figure}

To generate the integral image, all image data need to be loaded from global
memory to the shared memory. However, due to the limitation of
  the shared memory, the pixels in one row are divided into several segments
and integrated partly as shown in Fig.\ref{fig:Integration} (a). Considering the
limitations of data sharing between different GPU blocks, we only use one block
(Block0 in Fig.\ref{fig:Integration} (a)) for the integration of one row, which
means that each segment is integrated by the same block one by one (Round1,
Round2,...)  rather than processing multiple blocks in parallel.
In each block,
a two-layer parallel-integration strategy based on the {\itshape Kogge-Stone
  Adder} algorithm is used as shown in Fig.\ref{fig:Integration} (b) and (c),
since the threads work in the units of {\itshape warp}.
The {\itshape Kogge-Stone Adder} method is used because of its high
computational efficiency and suitability for thread-level parallel operations on
GPUs. It is not necessary to check the parity of the operand index for each
stage.
Figure.\ref{fig:Integration} (b) shows the parallel integration on the {\itshape
  Warp} layer. In Fig.\ref{fig:Integration} (b), ``$\circ$'' denotes the shared
memory used in the current segment. Before integration, all threads in each warp
are initialized with a variable $S1$, which represents the sum of the previous
segment (step 1). For the first segment in one row, $S1$ is initialized to
0. Then, each warp performs the integration independently and propagates its
intermediate results by shared memory according to the {\itshape Kogge-Stone
  Adder} method (step 2).  In this step, the result of each Warp $i$ is added to
Warp $i+2^{t-1}\ (i+2^{t-1}<N)$ through the shared memory, with $t$ representing
the number of repetitions increasing from 1 to $\lfloor{\log_2{N}\rfloor}$ and
$N$ representing the number of threads in each warp. Then, the sum of the
current segment is updated by the last thread (step 3); at the same time, the
integrated result of each thread is transferred to the global memory for the
integration along the $y$-axis.
\begin{figure}[t]
 \centerline{\includegraphics[width=3.2in]{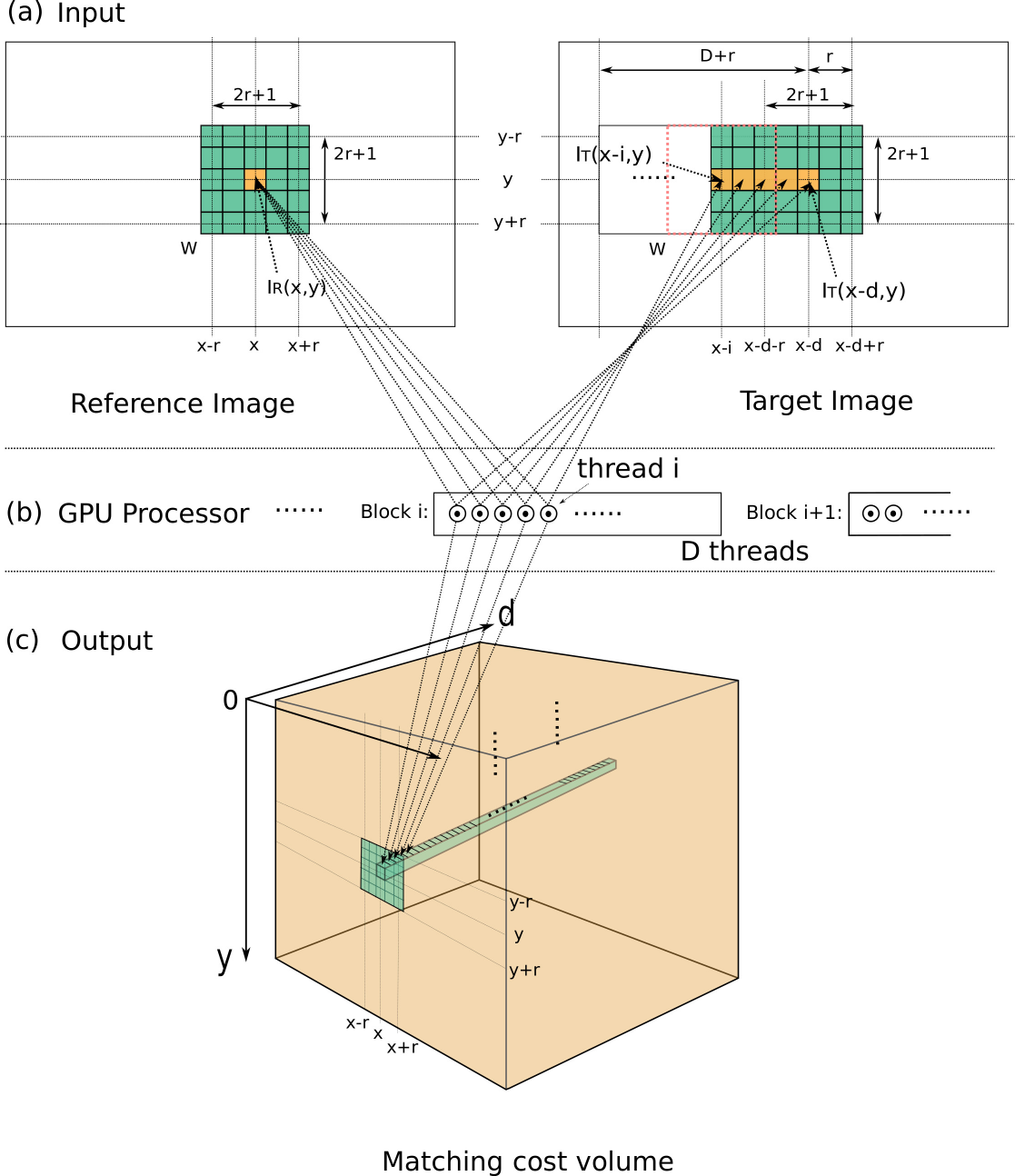}}
    \caption{Stereo matching on the GPU}
  \label{fig:StereoMatching}
\end{figure}
Figure.\ref{fig:Integration} (c) shows the parallel integration on the {\itshape
  Thread} layer. After the initialization shown in Fig.\ref{fig:Integration} (b)
step1, each thread loads the corresponding pixel value from the global memory to
the registers represented by ``$\bullet$'', and adds it to the initial value
$S1$. Then, integration is performed through the register shift among the
threads in the same warp using the method shown in Fig.\ref{fig:Integration} (b)
step 2, and the last result is stored in the shared memory.

By repeating the above steps until the integration of the last segment ends, the
integral image of each row can be calculated and used to obtain the entire
integral image along the $y$-axis.
Here, by transposing the matrix~\cite{r18}, integration can be transformed from
vertical to horizontal. However, this may be less effective than a direct
calculation when the vertical range is small, because the memory overhead
required by matrix transposition itself reduces the parallelism of the GPU
blocks. In this paper, we perform a sequential column-wise
  integration rather than using a parallel-computing method, because the height
of the image set is less than 400 pixels.
\subsubsection{Z$^2$-ZNCC on Stereo Matching}
\label{sec:Z$^2$-ZNCConstereomatching}
After the summations above, the terms in \eqref{equ:OP_ZNCC} can be easily
calculated with the exception of $\sum_{(x,y)\in
    W}{\Pi_{RT}(x,y,d)}$ is omitted in the following discussion to simplify the
description.
Here, we show that $\sum{\Pi_{RT}(x,y,d)}$ can be calculated efficiently by
scanning the image in a zigzag fashion.
Unlike the other summations, $\Pi_{RT}(x,y,d)$ represents a 3D-matching result
between reference pixels and multiple target pixels under different disparities.
Efficient calculation of $\sum{\Pi_{RT}(x,y,d)}$ is the most critical part of
our implementation.
\paragraph{Task Assignment}
\label{sec:taskassignment} 

As shown in Fig.\ref{fig:StereoMatching} (a), $I_R(x,y)$ is matched with $D$
pixels $I_T(x-d,y)$. To calculate $\sum{\Pi_{RT}(x,y,d)}$ for each $I_R(x,y)$,
$(2r+1)^2$ pixels around $I_R(x,y)$ and $(2r+1)\times{(D+2r)}$ pixels around
$I_T(x-d,y)\ (d\in[0,D))$ are required.
For this matching, one block is assigned because the pixel data loaded into the
shared memory can be reused to match adjacent pixels.
In each block, $D$ threads are assigned to perform the matching in parallel for
each corresponding $d$, as shown in Fig.\ref{fig:StereoMatching} (b).
Each thread $i$ calculates $\sum{\Pi_{RT}(x,y,i)}$ using the pixels in the
windows $W$ (centered at $I_R(x,y)$) and windows $W'$ (centered at
$I_T(x-i,y)$).
Here, $\Pi_{RT}(x,y,i)$ is calculated element by element, and their sum is
calculated efficiently via our approach described below.
After calculating $\sum{\Pi_{RT}(x,y,d)}$ (as shown in
Fig.\ref{fig:StereoMatching} (c)), the matching cost $C(x,y,d)$ can be
calculated according to \eqref{equ:OP_ZNCC} and then stored in the global memory
for use in the cost-aggregation stage.
With this task assignment, the data once loaded to the shared memory from the
global memory can be efficiently reused for the calculation of adjacent pixels
when $D$ is sufficiently large.
\paragraph{Zigzag Scanning}
\label{sec:4_2_2}

\begin{figure}[t]
 \centerline{\includegraphics[width=3in]{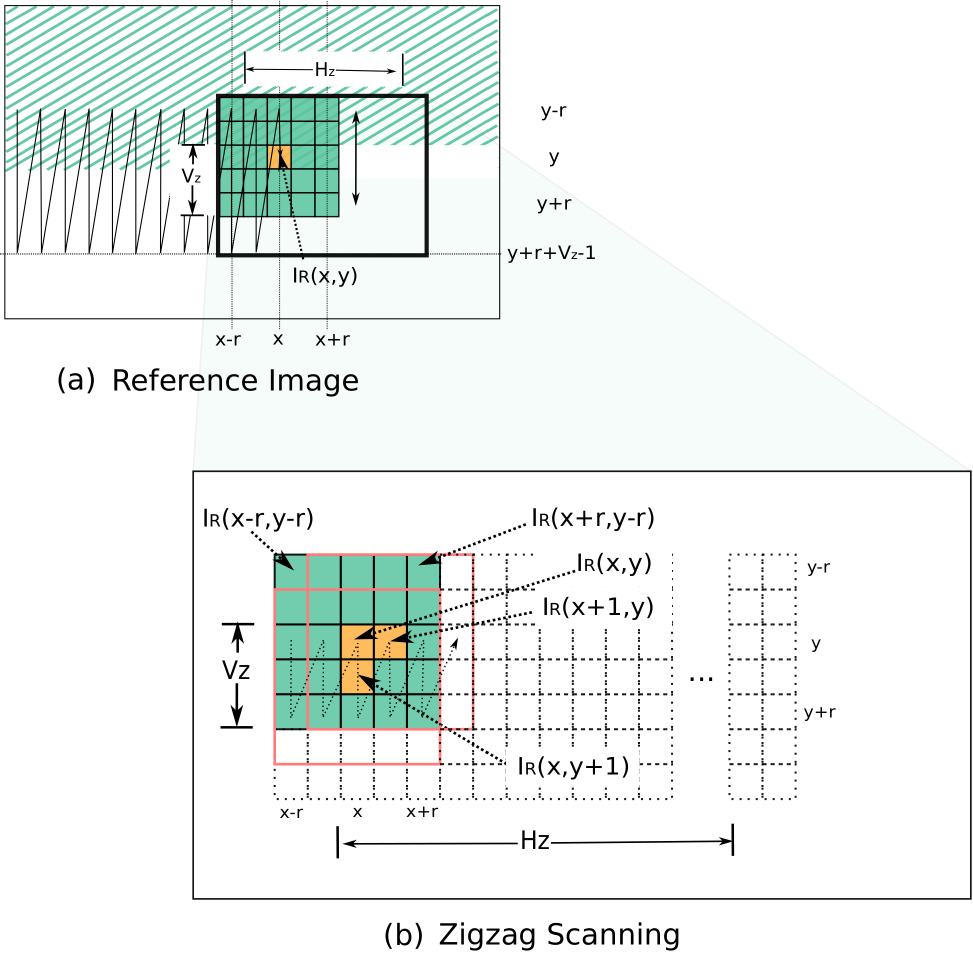}}
    \caption{Zigzag scanning}
    \label{fig:Zigzag}
    \vspace*{-0.4cm}
\end{figure}
In our approach, the image is scanned in a {\itshape zigzag} fashion, as shown
in Fig.\ref{fig:Zigzag} (a) along the $x$ and $y$ axis, while
$\sum{\Pi_{RT}(x,y,d)}$ is calculated for each pixel in parallel along the
$d$-axis.
$V_Z$ pixels in a column are processed first from top to bottom; then, the same
processing is repeated on the next column. This scanning method is repeated from
left to right.
In this zigzag scanning, the rows are segmented in the same way as in summation
Method 2, and
$(2r+V_Z)\times{(2r+H_Z)}$ pixels in the reference image and
$(2r+V_Z)\times{(2r+H_Z+D)}$ pixels in the target image are loaded into the
shared memory respectively as shown in Fig.\ref{fig:Zigzag} (b) (where $H_z$ is
a constant decided by the shared-memory size).
With this zigzag scanning, after $\sum{\Pi_{RT}(x,y,d)}$ was calculated,
$\sum{\Pi_{RT}(x,y+1,d)}$ and $\sum{\Pi_{RT}(x+1,y,d)}$ can be easily calculated
as follows:\vspace*{-0.3cm}

{\small
\begin{equation}
  \begin{split}
    \hspace*{-0.3cm} \sum_{(x,y)\in
      W}{\hspace*{-0.2cm}\Pi_{RT}(x,y{\small+}1,d)} &{\small=} \displaystyle
    \sum_{\Delta{x}{\mbox{\scriptsize=}}-r}^{r}{\sum_{\Delta{y}{\mbox{\scriptsize=}}-r}^{r}{\Pi_{RT}(x{\small+}\Delta{x},y{\small+}\Delta{y},d)}}\hspace*{-0.5cm}\\ &{\small+}\displaystyle
    \sum_{\Delta{x}=-r}^{r}{\Pi_{RT}(x{\small+}\Delta{x},y{\small+}r{\small+}1,d)}\\ &{\small-}\displaystyle
    \sum_{\Delta{x}=-r}^{r}{\Pi_{RT}(x{\small+}\Delta{x},y{\small-}r,d)},\\
\end{split}
\end{equation}
} {\small
\begin{equation}
  \begin{split}
    \hspace*{-0.3cm} \sum_{(x,y)\in
      W}{\hspace*{-0.2cm}\Pi_{RT}(x{\small+}1,y,d)} &{\small=} \displaystyle
    \sum_{\Delta{x}{\mbox{\scriptsize=}}-r}^{r}{\sum_{\Delta{y}{\mbox{\scriptsize=}}-r}^{r}{\Pi_{RT}(x{\small+}\Delta{x},y{\small+}\Delta{y},d)}}\hspace*{-0.5cm}\\ &{\small+}\displaystyle
    \sum_{\Delta{y}=-r}^{r}{\Pi_{RT}(x{\small+}r{\small+}1,y{\small+}\Delta{y},d)}\\ &{\small-}\displaystyle
    \sum_{\Delta{y}=-r}^{r}{\Pi_{RT}(x{\small-}r,y{\small+}\Delta{y},d)}.\\
\end{split}
\end{equation}
}

\begin{figure}
  \centerline{\includegraphics[width=3.3in]{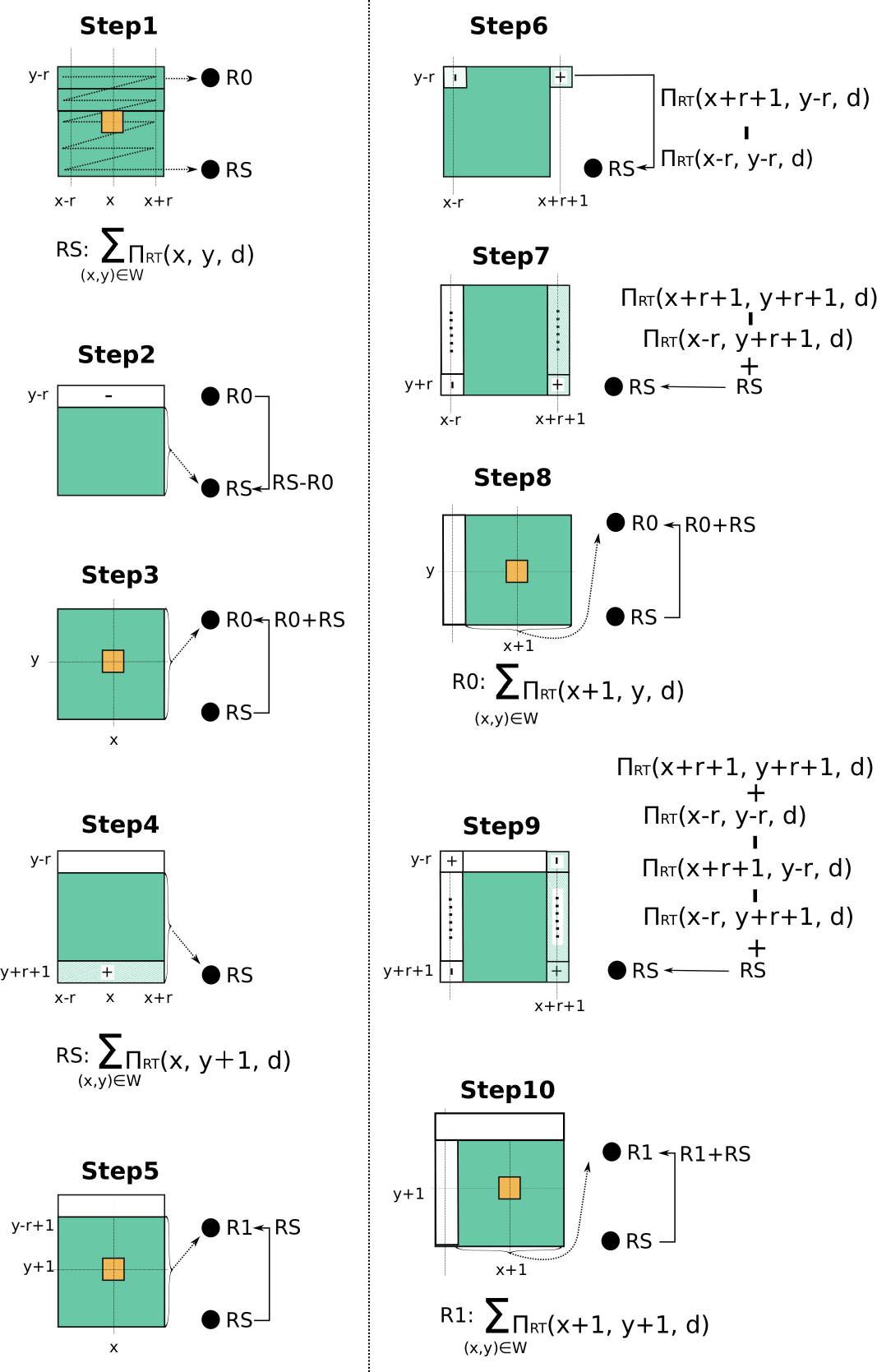}}
    \vspace*{-0.1cm}
    \caption{Efficient zigzag scanning}
    \label{fig:Z$^2$-ZNCC}
    \vspace*{-0.5cm}
\end{figure}

As shown in these two equations, the advantage of using the zigzag scanning
method is that as long as the sums of different rows and columns such as
$\sum_{\Delta{x}\in[-r,r]}$ ${\Pi_{RT}(x{\small+}\Delta{x}}$,${y{\small-}r,d)}$
and $\sum_{\Delta{y}\in[-r,r]}{\Pi_{RT}(x{\small-}r,y{\small+}\Delta{y},d)}$ can
be stored in the memory, they can be reused to efficiently calculate other sums
along both directions.
However, storing these intermediate results along the two directions requires a
huge number of registers, which may reduce the total efficiency.
\paragraph{Z$^2$-ZNCC}
\label{sec:Z$^2$-ZNCC}
To solve this problem, we propose a strategy for efficiently using
registers. Figure.\ref{fig:Z$^2$-ZNCC} shows the processing flow of the summation
of $(2r+1)^2$ pixel window. In this example, $V_Z=2$, meaning that two rows, $y$
and $y+1$, are processed during one zigzag scanning. The calculation process is
as follows:
  \begin{itemize}
  \item Step 1: $\sum{\Pi_{RT}(x,y,d)}$ is first calculated in order and stored
    in the register $RS$; then, it can be used to calculate the matching cost
    $C(x,y,d)$. During this step, in order to calculate
    $\sum{\Pi_{RT}(x,y+1,d)}$ efficiently, the sum of $2r+1$ pixels on row $y-r$
    is stored in the register $R0$.
  \item Step 2: The difference between $RS$ and $R0$ is calculated and stored in
    $RS$ to calculate $\sum{\Pi_{RT}(x,y+1,d)}$.
  \item Step 3: $\sum{\Pi_{RT}(x,y,d)}$ is still necessary for calculating
    $\sum{\Pi_{RT}(x+1,y,d)}$, but its value of $RS$ was discarded in Step2. On
    the other hand, the sum on row $y-r$ in $R0$ is no longer necessary. Thus,
    the difference stored in $RS$ is added back to $R0$ to recalculate
    $\sum{\Pi_{RT}(x,y,d)}$. This irregular procedure minimizes the number of
    registers used for this calculation and makes more threads active.
  \item Step 4: The sum of row $y+r+1$ is calculated and added to $RS$.  Then,
    $\sum{\Pi_{RT}(x,y+1,d)}$ is obtained and used to calculate the matching
    cost $C(x,y+1,d)$.
  \item Step 5: $\sum{\Pi_{RT}(x,y+1,d)}$ is stored in register $R1$ to calculate
    $\sum{\Pi_{RT}(x+1,y+1,d)}$ in the same way.
  \end{itemize}
At this point, $\sum{\Pi_{RT}(x,y,d)}$ and $\sum{\Pi_{RT}(x,y+1,d)}$ are stored
in $R0$ and $R1$ respectively, and these values are used to calculate
$\sum{\Pi_{RT}(x+1,y,d)}$ and $\sum{\Pi_{RT}(x+1,y+1,d)}$.
\begin{itemize}
\item Step 6,7: To calculate $\sum{\Pi_{RT}(x+1,y,d)}$ from
  $\sum{\Pi_{RT}(x,y,d)}$ in the same way, the sums of $2r+1$ pixels in columns
  $x-r$ and $x+r+1$ are required. In our implementation, to make more threads
  active by reducing the memory usage as much as possible, these sums are not
  stored in the memory during the above calculations. Then, the difference
  between the pixels in columns $x-r$ and $x+r+1$ is calculated and summed. In
  Step6, the difference of the uppermost pixels is calculated and stored in
  $RS$, and in Step 7, the differences of the other pixels are added to
  $RS$. Finally, $RS$ becomes the difference between $\sum{\Pi_{RT}(x,y,d)}$ and
  $\sum{\Pi_{RT}(x+1,y,d)}$.
\item Step 8: The accumulated difference is added to $R0$ and then
  $\sum{\Pi_{RT}(x+1,y,d)}$ can be obtained, and $C(x+1,y,d)$ is calculated.
\item Step 9: The difference stored in $RS$ is updated to calculate
  $\sum{\Pi_{RT}(x+1,y+1,d)}$ by adding and subtracting $\Pi_{RT}$ on the four
  corners.
\item Step 10: The accumulated difference $RS$ is added to $R1$ and then
  $\sum{\Pi_{RT}(x+1,y+1,d)}$ is obtained.
\end{itemize}
Using this method, we only need $V_Z+1$ registers for each thread to perform the
summation.
In our implementation, only the intermediate results along the $x$ axis are held
on registers, while those along the $y$ axis are recalculated.
This strategy is chosen because $V_Z$ is smaller than $H_Z$.
By repeating the above calculation continuously, the overall processing speed
can be greatly improved by limiting the number of registers for each thread, and
by making more threads active.

\subsection{Implementation of DT}
\label{sec:4_3}

\begin{figure}
 \center \includegraphics[width=3in]{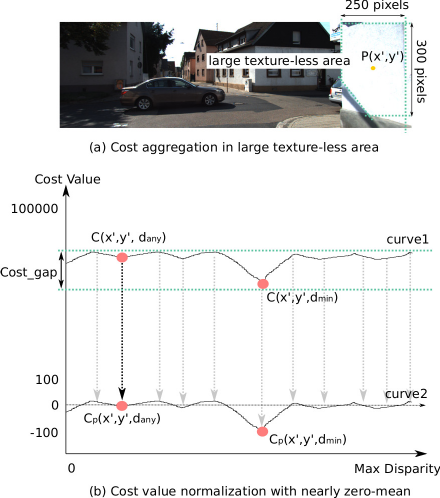}
    \caption{Cost-value normalization with nearly zero-mean}
  \label{fig:cost_penalizing}
\end{figure}

Since DT is performed based on the ZNCC computation results, the task assignment
is the same as that of ZNCC shown in Fig.\ref{fig:StereoMatching}.
In DT, because the cost values must be aggregated in four directions (left to
right, right to left, up to down, and down to up), a large memory space is
required.
The size of on-chip shared memory is too small for this purpose, and we
need to use the off-chip global memory.
Here, because both the ZNCC (\eqref{equ:ZNCC}) and weighting operations
($\eqref{equ:k1}$ to ${\eqref{equ:k4}}$) usually generate floating-point cost
values (32 bits), there is not only causes a great burden on data transmission,
but also a greater requirement for requires more shared-memory space, which
reduces the parallelism of multi-thread processing.
To solve these problems, it is usually good to use shorter integers (16 bits or
even 8 bits) to represent the cost values instead of the floating-point data
type.
However, a serious problem needs to be addressed here.
The magnitude of the aggregated cost values around the boundary of each area
decreases due to the use of weight. Hence, the magnitude is sufficiently small
for 16-bit integers but not for textureless regions.
Figure.\ref{fig:cost_penalizing} (a) shows an example of cost aggregation in a
large white wall with less texture. Let $P(x',y')$ be a pixel that belongs to
the wall in the image; its cost values $C(x',y',d)$ are accumulated from all
pixels in this area according to \eqref{equ:DT1} to \eqref{equ:DT4}.
Due to the size of the wall (around 250x300 pixels) and the range of the ZNCC
result ([0,1] as mentioned in Section~\ref{sec:3_1}), the accumulated
floating-point cost values $C(x',y',d)$ will largely exceed the upper limit on a
16-bit integer, causing overflow. As such, these cost values cannot be simply
converted to integers by multiplying by a coefficient.
In~\cite{r13}, we propose a solution to this problem by shifting the matching
cost of census to quickly reduce the value range and compressing the 16-bit data
into an 8-bit data with a 1-bit flag code. However, the shifting method is not
suitable for ZNCC because of its decimal cost value; furthermore, a 1-bit flag
code can only specify two positions on a 16-bit integer, which may reduce
accuracy.
Therefore, we upgrade the original solution by using a two-step strategy to
reduce the aggregated costs' data width and burden of transmission:
\begin{itemize}
  \item Use a cost-value normalization with a nearly zero-mean to represent the
    original floating-point cost values with 16-bit short integers.
  \item Apply a data encoding \& decoding method to further replace the
    normalized 16-bit short integers with 8-bit by using a 2-bit flag code.
\end{itemize}
The details of this strategy are as follows:
\paragraph{Cost-value Normalization with Nearly Zero-Mean}
\label{sec:4_3_2}
Figure.\ref{fig:cost_penalizing} (b) shows the change in the cost values of
$P(x',y')$ along the disparity (from 0 to $D-1$).
{\itshape Curve1} represents the change of $C(x',y',d)$ obtained by
\eqref{equ:DT4} (where $C_D(x',y',d)$ is used as final $C(x',y',d)$ as described
above), and $C(x',y',d_{min})$ shows the minimum value along the curve.
The corresponding disparity $d_{min}$ is the result obtained based on the
magnitude relationship of $C(x',y',d)$ according to \eqref{equ:WTA1}.
Therefore, as long as the magnitude relationship remains unchanged, changing the
values of $C(x',y',d)$ will not affect obtaining the correct
$d_{min}$.
Additionally, since the range of $C(x',y',d)$ is narrow (as shown in
Fig.\ref{fig:cost_penalizing}, {\itshape Cost\_{gap}}, the difference between
the max and min of $C(x',y',d)$ is much smaller than the values of $C(x',y',d)$
themselves. Since the range of {\itshape Cost\_{gap}} is narrow, each cost value
can be nearly zero-mean normalized by subtracting $C(x',y',d_{arb})$, where
$d_{arb}$ is an arbitrary disparity ($d_{arb}\neq{d_{min}}$).
By subtracting the median value, the range of $C(x',y',d)$ can be minimized; in
our implementation, however, an arbitrary value $C(x',y',d_{arb})$ is used to
simplify the calculation.
Then \eqref{equ:DT4} can be changed to:
\begin{align}
C'_{D}(x,y,d) &= C'_{U}(x,y,d) + C'_{D}(x,y+1,d)\cdot{W_D(x,y)} \nonumber\\ &-
C'_{U}(x,y,d_{arb}),
\label{equ:DT_new}
\end{align}
where $C'$ represents the normalized cost value such that the mean approaches
zero.
This effectively suppresses the increase in the aggregated cost values.
To further ensure that these values will not cause an overflow, the cost-value
normalization is extended in all directions. At the same time, each accumulated
floating-point cost value is scaled up to a 16-bit signed integer via an integer
coefficient $T$.
The value of $T$ needs to be carefully determined according to the actual
situation. The larger the value, the higher the accuracy but also the greater
the risk of overflow.
This approach not only effectively reduces the burden of data transmission, but
facilitates the further reduction of data width using the following method.

\begin{figure}
 \centering \includegraphics[width=2.7in]{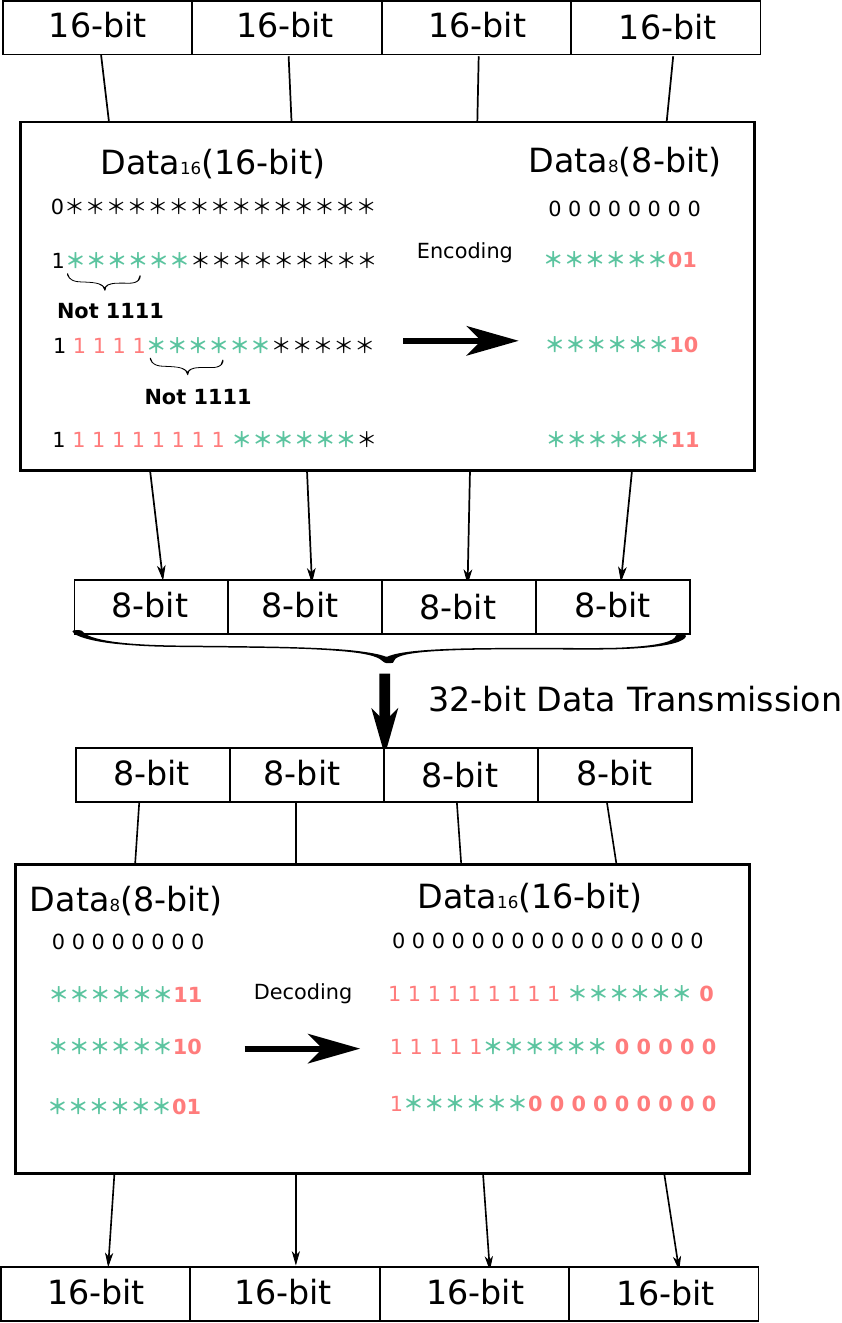}
    \caption{Data encoding \& decoding}
  \label{fig:encode}
\end{figure}

\paragraph{Data Encoding \& Decoding}
\label{sec:4_3_3}
After cost normalization,
the value of $C'(x',y',d_{arb})$ is reduced to 0 and some values, including
$C'(x',y',d_{min})$, become negative (when $d_{arb}\neq{d_{min}}$).
%
Therefore, it is possible to further reduce the data width by focusing only upon
the negative values while ignoring the positive ones.
Our method includes two stages: encoding and decoding.
Figure.\ref{fig:encode} shows the process of our method.
In the encoding stage, the 16-bit cost value is first compressed into an 8-bit
code containing a 6-bit value and a 2-bit flag.
This 6-bit value represents the 6 significant bits of the 16-bit integer, and
the 2-bit flag shows its position.
In the decoding stage, each 8-bit code is decompressed into a 16-bit
approximation by putting the 6-bit value in the 16-bit integer on the position
specified by the 2-bit flag code.
Between the stages, four adjacent 8-bit codes are packed into a 32-bit
integer for more efficient transmission on a GPU.
The details of the process in Fig.\ref{fig:encode} can be
described as follows:
\begin{itemize}
\item Encoding
  \begin{enumerate}
  \item As mentioned above, only negative values are used. Therefore, all
    positive values are set to 0 by checking the sign bit.
  \item Since two's complement is used to represent the negative values, we
    first test the 4-bit {$Data_{16}[14:11]$} to find whether it is $'1111'$. If
    not, ($Data_{16}[14:9]$) is chosen as the 6-bit code and $'01'$ is attached
    to it as the 2-bit flag to show the position of the 6-bit code in the
    original 16-bit integer. Then, an 8-bit dataset, $Data_8$, is constructed
    from the 16-bit integer.
  \item If
    $Data_{16}$[14:11] is $'1111'$, the next 4-bit code
    ($Data_{16}[10:7]$) is checked in the same way. If it is not
    $'1111'$, $Data_{16}[10:5]$ is chosen as the 6-bit code and the flag
    code $'10'$ is attached to show its position.
  \item Finally, if $Data_{16}[14:11]$ and $Data_{16}[10:7]$ are both $'1111'$,
    $Data_{16}[6:1]$ is chosen as the 6-bit code (no checking is necessary) and
    the 2-bit flag code $'11'$ is attached.
  \end{enumerate}
\end{itemize}
\begin{itemize}
\item Decoding
  \begin{enumerate}
  \item We first check whether $Data_8$ is $'0'$ or not. If it is, $Data_{16}$
    is also set to $'0'$; if not, its flag code $Data_8[1:0]$ is checked.
  \item If the 2-bit flag code is $'11'$, the 6-bit code at $Data_8[7:2]$ is
    copied to $Data_{16}[6:1]$; if the flag code is $'10'$, the 6-bit code is
    copied to $Data_{16}[10:5]$; if the flag code is $'01'$, the 6-bit code is
    copied to $Data_{16}[14:9]$.
  \item Then, the bits on the left-hand side of the copied 6-bit code in
    $Data_{16}$ are set to $'1'$ and the bits on the right side are set to
    $'0'$.
  \end{enumerate}
\end{itemize}
This method uses the 2-bit flag code, which specifies four position; and the
position of the 6-bit code is not continuous on a 16-bit integer. While this is
more accurate than the 1-bit flag code which specifies two positions, our
approach still loses more information than general 16-bit to 6-bit data-width
reduction.
However, according to our experiments, high speed processing is possible without
losing too much accuracy.

\section{Evaluation}
\label{sec:5}
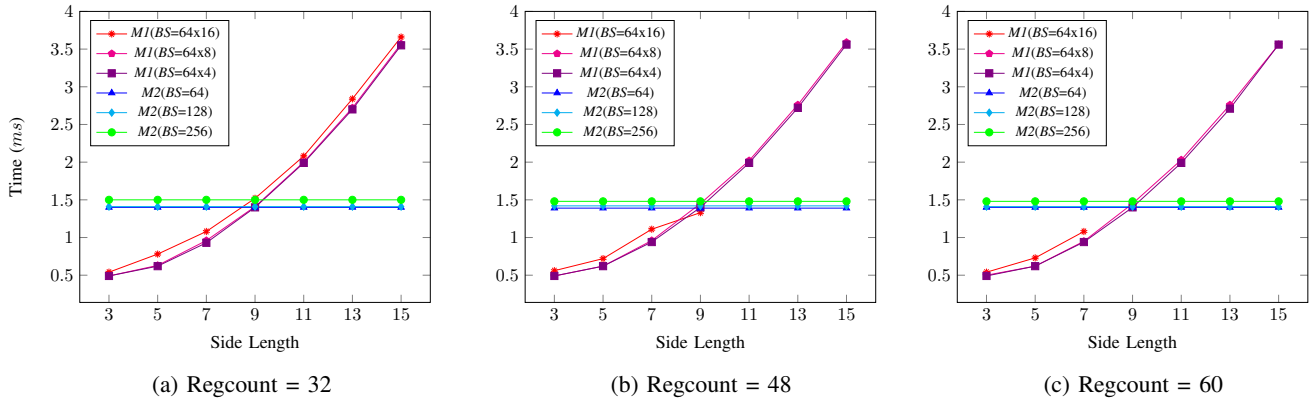
\begin{figure*}
\center
\begin{subfigure}{0.35\linewidth}
\begin{tikzpicture}[scale=0.68]
\begin{axis}
[ xlabel= Side Length, 
  ylabel= Time (${ms}$),
  legend style={font=\footnotesize},
legend entries ={{\itshape M1}({\itshape BS}=64x16), {\itshape M1}({\itshape BS}=64x8),{\itshape M1}({\itshape BS}=64x4),{\itshape M2}({\itshape BS}=64), {\itshape M2}({\itshape BS}=128), {\itshape M2}({\itshape BS}=256)},
legend pos = north west,
ymax = 4,
xtick={3,5,7,9,11,13,15},
ytick={0.5,1,1.5,2,2.5,3,3.5,4}
] 
\addplot[color=red,mark=10-pointed star] coordinates 
{ (	3	,	0.54	)
(	5	,	0.78	)
(	7	,	1.08	)
(	9	,	1.52	)
(	11	,	2.08	)
(	13	,	2.84	)
(	15	,	3.66	)

}; 
\addplot[color=magenta,mark=pentagon*] coordinates 
{ (	3	,	0.49	)
(	5	,	0.63	)
(	7	,	0.96	)
(	9	,	1.41	)
(	11	,	2	)
(	13	,	2.72	)
(	15	,	3.57	)

}; 
\addplot[color=violet,mark=square*] coordinates 
{ (	3	,	0.49	)
(	5	,	0.62	)
(	7	,	0.93	)
(	9	,	1.4	)
(	11	,	1.99	)
(	13	,	2.7	)
(	15	,	3.55	)
}; 
\addplot[color=blue,mark=triangle*] coordinates 
{ (	3	,	1.4	)
(	5	,	1.4	)
(	7	,	1.4	)
(	9	,	1.4	)
(	11	,	1.4	)
(	13	,	1.4	)
(	15	,	1.4	)

}; 
\addplot[color=cyan,mark=diamond*] coordinates 
{ (	3	,	1.41	)
(	5	,	1.41	)
(	7	,	1.41	)
(	9	,	1.41	)
(	11	,	1.41	)
(	13	,	1.41	)
(	15	,	1.41	)

}; 
\addplot[color=green,mark=*] coordinates 
{ (	3	,	1.5	)
(	5	,	1.5	)
(	7	,	1.5	)
(	9	,	1.5	)
(	11	,	1.5	)
(	13	,	1.5	)
(	15	,	1.5	)

}; 

\end{axis}
\end{tikzpicture}
\caption{Regcount = 32}
\end{subfigure}
\begin{subfigure}{0.31\linewidth}
\begin{tikzpicture}[scale=0.68]
\begin{axis}
[ xlabel= Side Length, 
  legend style={font=\footnotesize},
legend entries ={{\itshape M1}({\itshape BS}=64x16), {\itshape M1}({\itshape BS}=64x8),{\itshape M1}({\itshape BS}=64x4),{\itshape M2}({\itshape BS}=64), {\itshape M2}({\itshape BS}=128), {\itshape M2}({\itshape BS}=256)},
legend pos = north west,
ymax = 4,
xtick={3,5,7,9,11,13,15},
ytick={0.5,1,1.5,2,2.5,3,3.5,4}
] 
\addplot[color=red,mark=10-pointed star] coordinates 
{ (	3	,	0.56	)
(	5	,	0.72	)
(	7	,	1.11	)
(	9	,	1.330	)
}; 
\addplot[color=magenta,mark=pentagon*] coordinates 
{ (	3	,	0.49	)
(	5	,	0.62	)
(	7	,	0.96	)
(	9	,	1.45	)
(	11	,	2.02	)
(	13	,	2.76	)
(	15	,	3.59	)

}; 
\addplot[color=violet,mark=square*] coordinates 
{ (	3	,	0.49	)
(	5	,	0.62	)
(	7	,	0.94	)
(	9	,	1.4	)
(	11	,	1.99	)
(	13	,	2.72	)
(	15	,	3.56    )
}; 
\addplot[color=blue,mark=triangle*] coordinates 
{ (	3	,	1.39	)
(	5	,	1.39	)
(	7	,	1.39	)
(	9	,	1.39	)
(	11	,	1.39	)
(	13	,	1.39	)
(	15	,	1.39	)

}; 
\addplot[color=cyan,mark=diamond*] coordinates 
{ (	3	,	1.42	)
(	5	,	1.42	)
(	7	,	1.42	)
(	9	,	1.42	)
(	11	,	1.42	)
(	13	,	1.42	)
(	15	,	1.42	)

}; 
\addplot[color=green,mark=*] coordinates 
{ (	3	,	1.48	)
(	5	,	1.48	)
(	7	,	1.48	)
(	9	,	1.48	)
(	11	,	1.48	)
(	13	,	1.48	)
(	15	,	1.48	)

}; 

\end{axis}
\end{tikzpicture}
\caption{Regcount = 48}
\end{subfigure}
\begin{subfigure}{0.31\linewidth}
\begin{tikzpicture}[scale=0.68]
\begin{axis}
[ xlabel= Side Length, 
  legend style={font=\footnotesize},
legend entries ={{\itshape M1}({\itshape BS}=64x16), {\itshape M1}({\itshape BS}=64x8),{\itshape M1}({\itshape BS}=64x4),{\itshape M2}({\itshape BS}=64), {\itshape M2}({\itshape BS}=128), {\itshape M2}({\itshape BS}=256)},
legend pos = north west,
ymax = 4,
xtick={3,5,7,9,11,13,15},
ytick={0.5,1,1.5,2,2.5,3,3.5,4}
] 
\addplot[color=red,mark=10-pointed star] coordinates 
{ (	3	,	0.54	)
(	5	,	0.73	)
(	7	,	1.08	)

}; 
\addplot[color=magenta,mark=pentagon*] coordinates 
{ (	3	,	0.5	)
(	5	,	0.62	)
(	7	,	0.95	)
(	9	,	1.45	)
(	11	,	2.03	)
(	13	,	2.76	)
(	15	,	3.56	)

}; 
\addplot[color=violet,mark=square*] coordinates 
{ (	3	,	0.49	)
(	5	,	0.62	)
(	7	,	0.94	)
(	9	,	1.4	)
(	11	,	1.99	)
(	13	,	2.71    )
(	15	,	3.56	)
}; 
 
\addplot[color=blue,mark=triangle*] coordinates 
{ (	3	,	1.40	)
(	5	,	1.4	)
(	7	,	1.4	)
(	9	,	1.4	)
(	11	,	1.4	)
(	13	,	1.4	)
(	15	,	1.4	)

}; 
\addplot[color=cyan,mark=diamond*] coordinates 
{ (	3	,	1.41	)
(	5	,	1.41	)
(	7	,	1.41	)
(	9	,	1.41	)
(	11	,	1.41	)
(	13	,	1.41	)
(	15	,	1.41	)

}; 
\addplot[color=green,mark=*] coordinates 
{ (	3	,	1.48	)
(	5	,	1.48	)
(	7	,	1.48	)
(	9	,	1.48	)
(	11	,	1.48	)
(	13	,	1.48	)
(	15	,	1.48	)

}; 

\end{axis}
\end{tikzpicture}
\caption{Regcount = 60}
\end{subfigure}
\caption{Processing speed comparison for summation}
\label{fig:summation_compare}
\end{figure*}

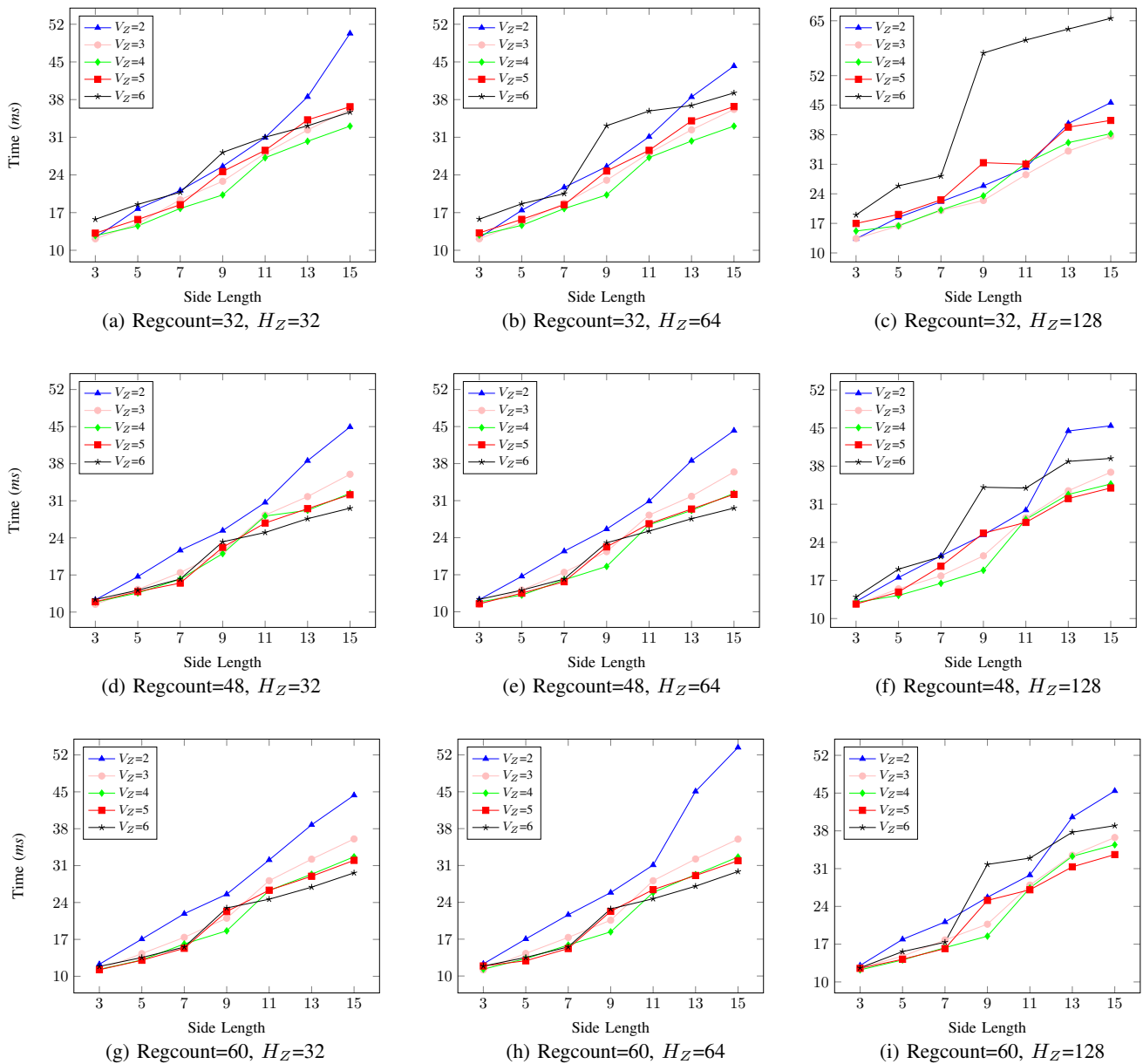
\begin{figure*}
\center
\begin{subfigure}{0.35\linewidth}
\begin{tikzpicture}[scale=0.68]
\begin{axis}
[ xlabel= Side Length, 
ylabel= Time ({\itshape ms}),
legend style={font=\footnotesize},
legend entries ={$V_{Z}$=2, $V_{Z}$=3, $V_{Z}$=4, $V_{Z}$=5, $V_{Z}$=6},
legend pos = north west,
ymax = 55,
xtick={3,5,7,9,11,13,15},
ytick={10,17,24,31,38,45,52}
] 
\addplot[color=blue,mark=triangle*] coordinates 
{ (	3	,	12.35	)
(	5	,	17.70	)
(	7	,	21.11	)
(	9	,	25.6	)
(	11	,	30.94    )
(	13	,	38.51	)
(	15	,	50.29   )
}; 
\addplot[color=pink,mark=*] coordinates 
{ (	3	,	12.13	)
(	5	,	14.95	)
(	7	,	19.39	)
(	9	,	22.82	)
(	11	,	28.02	)
(	13	,	32.36	)
(	15	,	36.17	)
}; 
\addplot[color=green,mark=diamond*] coordinates 
{ (	3	,	12.71	)
(	5	,	14.53	)
(	7	,	17.77	)
(	9	,	20.28	)
(	11	,	27.19	)
(	13	,	30.25	)
(	15	,	33.07	)
};

\addplot[color=red,mark=square*] coordinates 
{ (	3	,	13.18	)
(	5	,	15.72	)
(	7	,	18.47	)
(	9	,	24.63	)
(	11	,	28.57	)
(	13	,	34.23	)
(	15	,	36.67   )
};

\addplot[color=black,mark=star] coordinates 
{ (	3	,	15.75	)
(	5	,	18.52	)
(	7	,	20.78	)
(	9	,	28.19	)
(	11	,	31.01	)
(	13	,	33.12	)
(	15	,	35.68   )
};       
\end{axis}
\end{tikzpicture}
\vspace*{-0.2cm}
\caption{Regcount=32, $H_Z$=32}
\end{subfigure}
\vspace*{0.2cm}
\begin{subfigure}{0.31\linewidth}
\begin{tikzpicture}[scale=0.68]
\begin{axis}
[ xlabel= Side Length, 
legend style={font=\footnotesize},
legend entries ={$V_{Z}$=2, $V_{Z}$=3, $V_{Z}$=4, $V_{Z}$=5, $V_{Z}$=6},
legend pos = north west,
ymax = 55,
xtick={3,5,7,9,11,13,15},
ytick={10,17,24,31,38,45,52}
] 
\addplot[color=blue,mark=triangle*] coordinates 
{ (	3	,	12.33	)
(	5	,	17.43	)
(	7	,	21.68	)
(	9	,	25.55	)
(	11	,	31.09    )
(	13	,	38.49	)
(	15	,	44.23   )
}; 
\addplot[color=pink,mark=*] coordinates 
{ (	3	,	12.13	)
(	5	,	15.18	)
(	7	,	18.73	)
(	9	,	23.02	)
(	11	,	28.06	)
(	13	,	32.38	)
(	15	,	36.18	)
}; 
\addplot[color=green,mark=diamond*] coordinates 
{ (	3	,	12.81	)
(	5	,	14.63	)
(	7	,	17.74	)
(	9	,	20.29	)
(	11	,	27.23	)
(	13	,	30.30	)
(	15	,	33.07	)
};

\addplot[color=red,mark=square*] coordinates 
{ (	3	,	13.23	)
(	5	,	15.74	)
(	7	,	18.46	)
(	9	,	24.72	)
(	11	,	28.57	)
(	13	,	34.05	)
(	15	,	36.71   )
};

\addplot[color=black,mark=star] coordinates 
{ (	3	,	15.78	)
(	5	,	18.64	)
(	7	,	20.53	)
(	9	,	33.14	)
(	11	,	35.88	)
(	13	,	36.9	)
(	15	,	39.24   )
};       
\end{axis}
\end{tikzpicture}
\vspace*{-0.2cm}
\caption{Regcount=32, $H_Z$=64 }
\end{subfigure}
\vspace*{0.2cm}
\begin{subfigure}{0.31\linewidth}
\begin{tikzpicture}[scale=0.68]
\begin{axis}
[ xlabel= Side Length,  
legend style={font=\footnotesize},
legend entries ={$V_{Z}$=2, $V_{Z}$=3, $V_{Z}$=4, $V_{Z}$=5, $V_{Z}$=6},
legend pos = north west,
ymax = 68,
xtick={3,5,7,9,11,13,15},
ytick={10,17,24,31,38,45,52,65}
] 
\addplot[color=blue,mark=triangle*] coordinates 
{ (	3	,	13.34	)
(	5	,	18.42	)
(	7	,	22.13	)
(	9	,	25.89	)
(	11	,	30.19    )
(	13	,	40.65	)
(	15	,	45.6   )
}; 
\addplot[color=pink,mark=*] coordinates 
{ (	3	,	13.41	)
(	5	,	16.38	)
(	7	,	20.04	)
(	9	,	22.42	)
(	11	,	28.53	)
(	13	,	34.13	)
(	15	,	37.66	)
}; 
\addplot[color=green,mark=diamond*] coordinates 
{ (	3	,	15.19	)
(	5	,	16.42	)
(	7	,	20.17	)
(	9	,	23.53	)
(	11	,	31.26	)
(	13	,	36.13	)
(	15	,	38.24	)
};

\addplot[color=red,mark=square*] coordinates 
{ (	3	,	16.98	)
(	5	,	19.12	)
(	7	,	22.56	)
(	9	,	31.37	)
(	11	,	31.01	)
(	13	,	39.8	)
(	15	,	41.39   )
};

\addplot[color=black,mark=star] coordinates 
{ (	3	,	19	)
(	5	,	25.88	)
(	7	,	28.21	)
(	9	,	57.35	)
(	11	,	60.43	)
(	13	,	63.03	)
(	15	,	65.57   )
};       
\end{axis}
\end{tikzpicture}
\vspace*{-0.2cm}
\caption{Regcount=32, $H_Z$=128}
\end{subfigure}
\vspace*{0.2cm}

\begin{subfigure}{0.35\linewidth}
\begin{tikzpicture}[scale=0.68]
\begin{axis}
[ xlabel= Side Length, 
ylabel= Time ({\itshape ms}),
legend style={font=\footnotesize},
legend entries ={$V_{Z}$=2, $V_{Z}$=3, $V_{Z}$=4, $V_{Z}$=5, $V_{Z}$=6},
legend pos = north west,
ymax = 55,
xtick={3,5,7,9,11,13,15},
ytick={10,17,24,31,38,45,52}
] 
\addplot[color=blue,mark=triangle*] coordinates 
{ (	3	,	12.29	)
(	5	,	16.69	)
(	7	,	21.64	)
(	9	,	25.38	)
(	11	,	30.68    )
(	13	,	38.56	)
(	15	,	44.94   )
}; 
\addplot[color=pink,mark=*] coordinates 
{ (	3	,	11.46	)
(	5	,	14.27	)
(	7	,	17.44	)
(	9	,	21.43	)
(	11	,	28.30	)
(	13	,	31.79	)
(	15	,	36.01	)
}; 
\addplot[color=green,mark=diamond*] coordinates 
{ (	3	,	11.89	)
(	5	,	13.61	)
(	7	,	16.22	)
(	9	,	21.03	)
(	11	,	28.13	)
(	13	,	29.20	)
(	15	,	32.41	)
};

\addplot[color=red,mark=square*] coordinates 
{ (	3	,	11.95	)
(	5	,	13.85	)
(	7	,	15.47	)
(	9	,	22.22	)
(	11	,	26.80	)
(	13	,	29.52	)
(	15	,	32.14   )
};

\addplot[color=black,mark=star] coordinates 
{ (	3	,	12.37	)
(	5	,	14.12	)
(	7	,	16.23	)
(	9	,	23.26	)
(	11	,	25.02	)
(	13	,	27.62	)
(	15	,	29.60   )
};       
\end{axis}
\end{tikzpicture}
\vspace*{-0.2cm}
\caption{Regcount=48, $H_Z$=32}
\end{subfigure}
\vspace*{0.2cm}
\begin{subfigure}{0.31\linewidth}
\begin{tikzpicture}[scale=0.68]
\begin{axis}
[ xlabel= Side Length, 
legend style={font=\footnotesize},
legend entries ={$V_{Z}$=2, $V_{Z}$=3, $V_{Z}$=4, $V_{Z}$=5, $V_{Z}$=6},
legend pos = north west,
ymax = 55,
xtick={3,5,7,9,11,13,15},
ytick={10,17,24,31,38,45,52}
] 
\addplot[color=blue,mark=triangle*] coordinates 
{ (	3	,	12.30	)
(	5	,	16.7	)
(	7	,	21.44	)
(	9	,	25.63	)
(	11	,	30.88    )
(	13	,	38.55	)
(	15	,	44.23   )
}; 
\addplot[color=pink,mark=*] coordinates 
{ (	3	,	11.41	)
(	5	,	14.11	)
(	7	,	17.46	)
(	9	,	21.36	)
(	11	,	28.27	)
(	13	,	31.82	)
(	15	,	36.42	)
}; 
\addplot[color=green,mark=diamond*] coordinates 
{ (	3	,	11.89	)
(	5	,	13.15	)
(	7	,	16.04	)
(	9	,	18.57	)
(	11	,	26.43	)
(	13	,	29.16	)
(	15	,	32.38	)
};

\addplot[color=red,mark=square*] coordinates 
{ (	3	,	11.52	)
(	5	,	13.52	)
(	7	,	15.68	)
(	9	,	22.26	)
(	11	,	26.64	)
(	13	,	29.42	)
(	15	,	32.18   )
};

\addplot[color=black,mark=star] coordinates 
{ (	3	,	12.35	)
(	5	,	14.09	)
(	7	,	16.20	)
(	9	,	23.04	)
(	11	,	25.25	)
(	13	,	27.58	)
(	15	,	29.59   )
};       
\end{axis}
\end{tikzpicture}
\vspace*{-0.2cm}
\caption{Regcount=48, $H_Z$=64 }
\end{subfigure}
\vspace*{0.2cm}
\begin{subfigure}{0.31\linewidth}
\begin{tikzpicture}[scale=0.68]
\begin{axis}
[ xlabel= Side Length, 
legend style={font=\footnotesize},
legend entries ={$V_{Z}$=2, $V_{Z}$=3, $V_{Z}$=4, $V_{Z}$=5, $V_{Z}$=6},
legend pos = north west,
ymax = 55,
xtick={3,5,7,9,11,13,15},
ytick={10,17,24,31,38,45,52}
] 
\addplot[color=blue,mark=triangle*] coordinates 
{ (	3	,	13.12	)
(	5	,	17.53	)
(	7	,	21.52	)
(	9	,	25.36	)
(	11	,	29.91    )
(	13	,	44.46	)
(	15	,	45.41   )
}; 
\addplot[color=pink,mark=*] coordinates 
{ (	3	,	12.61	)
(	5	,	15.45	)
(	7	,	17.84	)
(	9	,	21.52	)
(	11	,	28.49	)
(	13	,	33.48	)
(	15	,	36.88	)
}; 
\addplot[color=green,mark=diamond*] coordinates 
{ (	3	,	12.99	)
(	5	,	14.24	)
(	7	,	16.47	)
(	9	,	18.87	)
(	11	,	28.22	)
(	13	,	32.8	)
(	15	,	34.72	)
};

\addplot[color=red,mark=square*] coordinates 
{ (	3	,	12.67	)
(	5	,	14.87	)
(	7	,	19.61	)
(	9	,	25.68	)
(	11	,	27.66	)
(	13	,	32.04	)
(	15	,	33.99   )
};

\addplot[color=black,mark=star] coordinates 
{ (	3	,	13.96	)
(	5	,	19.07	)
(	7	,	21.38	)
(	9	,	34.12	)
(	11	,	33.97	)
(	13	,	38.87	)
(	15	,	39.42   )
};       
\end{axis}
\end{tikzpicture}
\vspace*{-0.2cm}
\caption{Regcount=48, $H_Z$=128}
\end{subfigure}
\vspace*{0.2cm}

\begin{subfigure}{0.35\linewidth}
\begin{tikzpicture}[scale=0.68]
\begin{axis}
[ xlabel= Side Length, 
ylabel= Time ({\itshape ms}),
legend style={font=\footnotesize},
legend entries ={$V_{Z}$=2, $V_{Z}$=3, $V_{Z}$=4, $V_{Z}$=5, $V_{Z}$=6},
legend pos = north west,
ymax = 55,
xtick={3,5,7,9,11,13,15},
ytick={10,17,24,31,38,45,52}
] 
\addplot[color=blue,mark=triangle*] coordinates 
{ (	3	,	12.28	)
(	5	,	17.06	)
(	7	,	21.87	)
(	9	,	25.57	)
(	11	,	32.07    )
(	13	,	38.74	)
(	15	,	44.35   )
}; 
\addplot[color=pink,mark=*] coordinates 
{ (	3	,	11.22	)
(	5	,	14.30	)
(	7	,	17.39	)
(	9	,	21.02	)
(	11	,	28.14	)
(	13	,	32.22	)
(	15	,	36.06	)
}; 
\addplot[color=green,mark=diamond*] coordinates 
{ (	3	,	11.29	)
(	5	,	13.10	)
(	7	,	16.13	)
(	9	,	18.62	)
(	11	,	26.29	)
(	13	,	29.36	)
(	15	,	32.67	)
};

\addplot[color=red,mark=square*] coordinates 
{ (	3	,	11.24	)
(	5	,	13.04	)
(	7	,	15.31	)
(	9	,	22.29	)
(	11	,	26.33	)
(	13	,	28.98	)
(	15	,	32.00   )
};

\addplot[color=black,mark=star] coordinates 
{ (	3	,	11.87	)
(	5	,	13.56	)
(	7	,	15.54	)
(	9	,	22.96	)
(	11	,	24.62	)
(	13	,	26.91	)
(	15	,	29.62   )
};       
\end{axis}
\end{tikzpicture}
\vspace*{-0.2cm}
\caption{Regcount=60, $H_Z$=32}
\end{subfigure}
\begin{subfigure}{0.31\linewidth}
\begin{tikzpicture}[scale=0.68]
\begin{axis}
[ xlabel= Side Length, 
legend style={font=\footnotesize},
legend entries ={$V_{Z}$=2, $V_{Z}$=3, $V_{Z}$=4, $V_{Z}$=5, $V_{Z}$=6},
legend pos = north west,
ymax = 55,
xtick={3,5,7,9,11,13,15},
ytick={10,17,24,31,38,45,52}
] 
\addplot[color=blue,mark=triangle*] coordinates 
{ (	3	,	12.30	)
(	5	,	17.04	)
(	7	,	21.63	)
(	9	,	25.85	)
(	11	,	31.08    )
(	13	,	45.08	)
(	15	,	53.42   )
}; 
\addplot[color=pink,mark=*] coordinates 
{ (	3	,	11.17	)
(	5	,	14.29	)
(	7	,	17.33	)
(	9	,	20.6	)
(	11	,	28.10	)
(	13	,	32.23	)
(	15	,	36	)
}; 
\addplot[color=green,mark=diamond*] coordinates 
{ (	3	,	11.3	)
(	5	,	13.29	)
(	7	,	15.89	)
(	9	,	18.39	)
(	11	,	25.87	)
(	13	,	29.26	)
(	15	,	32.61	)
};

\addplot[color=red,mark=square*] coordinates 
{ (	3	,	11.91	)
(	5	,	12.89	)
(	7	,	15.22	)
(	9	,	22.28	)
(	11	,	26.41	)
(	13	,	29.09	)
(	15	,	31.9   )
};

\addplot[color=black,mark=star] coordinates 
{ (	3	,	11.89	)
(	5	,	13.53	)
(	7	,	15.53	)
(	9	,	22.75	)
(	11	,	24.68	)
(	13	,	27.06	)
(	15	,	29.85   )
};       
\end{axis}
\end{tikzpicture}
\vspace*{-0.2cm}
\caption{Regcount=60, $H_Z$=64 }
\end{subfigure}
\begin{subfigure}{0.31\linewidth}
\begin{tikzpicture}[scale=0.68]
\begin{axis}
[ xlabel= Side Length, 
legend style={font=\footnotesize},
legend entries ={$V_{Z}$=2, $V_{Z}$=3, $V_{Z}$=4, $V_{Z}$=5, $V_{Z}$=6},
legend pos = north west,
ymax = 55,
xtick={3,5,7,9,11,13,15},
ytick={10,17,24,31,38,45,52}
] 
\addplot[color=blue,mark=triangle*] coordinates 
{ (	3	,	13.05	)
(	5	,	17.89	)
(	7	,	21.11	)
(	9	,	25.69	)
(	11	,	29.81    )
(	13	,	40.54	)
(	15	,	45.4   )
}; 
\addplot[color=pink,mark=*] coordinates 
{ (	3	,	12.62	)
(	5	,	14.77	)
(	7	,	17.78	)
(	9	,	20.7	)
(	11	,	27.95	)
(	13	,	33.48	)
(	15	,	36.77	)
}; 
\addplot[color=green,mark=diamond*] coordinates 
{ (	3	,	12.24	)
(	5	,	14.07	)
(	7	,	16.35	)
(	9	,	18.48	)
(	11	,	27.42	)
(	13	,	33.29	)
(	15	,	35.42	)
};

\addplot[color=red,mark=square*] coordinates 
{ (	3	,	12.52	)
(	5	,	14.2	)
(	7	,	16.18	)
(	9	,	25.11	)
(	11	,	27.08	)
(	13	,	31.33	)
(	15	,	33.6   )
};

\addplot[color=black,mark=star] coordinates 
{ (	3	,	12.61	)
(	5	,	15.62	)
(	7	,	17.36	)
(	9	,	31.78	)
(	11	,	32.95	)
(	13	,       37.75	)
(	15	,	38.96   )
};       
\end{axis}
\end{tikzpicture}
\vspace*{-0.2cm}
\caption{Regcount=60, $H_Z$=128}
\end{subfigure}
\caption{Processing speed evaluation for Z$^2$-ZNCC}
\label{fig:ZNCC_evaluation}








\end{figure*}
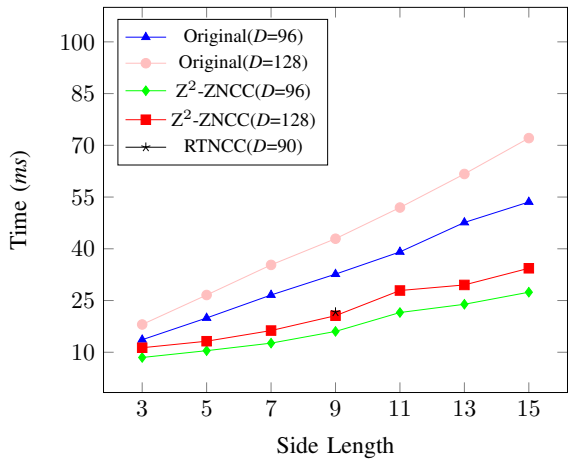
\begin{figure}
\center
\begin{tikzpicture}[scale=0.9]
\begin{axis}
[ xlabel= Side Length, 
ylabel= Time ({\itshape ms}),
legend style={font=\footnotesize},
legend entries ={Original({\itshape D}=96), Original({\itshape D}=128), Z$^2$-ZNCC({\itshape D}=96), Z$^2$-ZNCC({\itshape D}=128), RTNCC({\itshape D}=90)},
legend pos = north west,
ymax = 110,
xtick={3,5,7,9,11,13,15},
ytick={10,25,40,55,70,85,100}
] 
\addplot[color=blue,mark=triangle*] coordinates 
{ (	3	,	13.63	)
(	5	,	19.94	)
(	7	,	26.6	)
(	9	,	32.65	)
(	11	,	39.1    )
(	13	,	47.63	)
(	15	,	53.57   )
}; 
\addplot[color=pink,mark=*] coordinates 
{ (	3	,	18.05	)
(	5	,	26.6	)
(	7	,	35.33	)
(	9	,	42.92	)
(	11	,	51.96	)
(	13	,	61.69	)
(	15	,	72.09	)
}; 
\addplot[color=green,mark=diamond*] coordinates 
{ (	3	,	8.48	)
(	5	,	10.46	)
(	7	,	12.64	)
(	9	,	16.04	)
(	11	,	21.52	)
(	13	,	23.91	)
(	15	,	27.41	)
};

\addplot[color=red,mark=square*] coordinates 
{ (	3	,	11.31	)
(	5	,	13.2	)
(	7	,	16.29	)
(	9	,	20.6	)
(	11	,	27.9	)
(	13	,	29.52	)
(	15	,	34.35	)
};

\addplot[color=black,mark=star] coordinates 
{ 
(	9	,	21.645	)
};        
\end{axis}
\end{tikzpicture}
\caption{Processing speed comparison for Z$^2$-ZNCC}
\label{fig:ZNCC_compare}
\end{figure}

We implemented our acceleration approach on an embedded GPU Jetson Tx2 and
evaluated
\begin{enumerate}
\item summation,
\item Z$^2$-ZNCC,
\item FastDT, and
\item the processing speed and matching accuracy of the stereo-matching system
  based on the Z$^2$-ZNCC, census, semi-global matching (SGM), and FastDT,
\end{enumerate}
algorithms using the KITTI 2015~\cite{r6} benchmark.
\subsection{Evaluation of Summation}
\label{sec:5_1}
The processing speeds of the two summation methods described in Section
\ref{sec:4} are compared using 1,280x384 pixel images.
In our evaluation, the maximum number of registers ({\itshape Regcount}) for
each thread is limited to 32, 48, and 60; then comparisons are performed for
different window sizes and GPU blocks.

Figure.\ref{fig:summation_compare} compares the results of the two summation
methods.
In each graph, {\itshape M1} and {\itshape M2} represent the two methods and
{\itshape BS} represent the GPU block size. The $x$-axis represents matching
windows' side length from 3 to 15 and the $y$-axis represents their
corresponding processing times.
In Method 1, each block processes 64 columns and three sets of rows: 16, 8, and
4.
As the side length increases, the processing time increases accordingly. This is
occurs not only due to the increase in the amount of the calculation, but also
to the increase in memory occupancy, which reduces the number of active threads.
On the other hand, in Method 2, each block processes three sets of columns: 64,
128, and 256. Because the sum is calculated based on the integral image, the
processing time does not change with side length.
Here, we note that in all cases, the processing time does not change
significantly with the GPU-block size because the parallelism of threads is not
affected.
For all three different block sizes in Method 2, the processing times for the
first two steps of obtaining an integral image is about 355 $\mu{s}$ and 496
$\mu{s}$, respectively, and the total time including the averaging calculation
is close to 1.4 ms.
Due to the integration, the processing speed of Method 2 is not as fast as that
of Method 1 when the window size is smaller than 9x9.
Therefore, the methods can be chosen according to actual requirements.
However, when {\itshape BS} = 64x16 and {\itshape Regcount} = 48 (or {\itshape
  Regcount} = 60), Method 1 becomes invalid when the side length is larger than 9
(or 7);
this is because the large number of threads causes the number of allocated
registers to exceed the allowable upper limit.
Therefore, for Method 1, a small GPU block should be chosen so as to ensure that
the summation can be performed correctly.

\subsection{Evaluation of Z$^2$-ZNCC}
\label{sec:5_2}
Figure.\ref{fig:ZNCC_evaluation} shows the evaluation results for Z$^2$-ZNCC using
the same 1,280x384 size images with a maximum disparity of 128.
To show the performance of Z$^2$-ZNCC more clearly, Method 2 is used for the
summation because of its consistent processing speed.
In addition to {\itshape Regcount}, the evaluation is performed under various
$V_Z$ and $H_Z$ conditions, as described in Section.\ref{sec:4}.
$V_Z$ is changed from 2 to 6 and $H_Z$ is set to 32, 64, and 128. A large $V_Z$
means that the rate of data reuse is high; however, it also requires a larger
{\itshape Regcount}, which will affect the parallelism of threads.
Similarly, a large $H_Z$ suggests that a large number of $\Pi_{RT}(x,y,d)$ need
to be calculated at the same time, which also requires many registers.
According to this figure, we note that in most cases, when $V_Z$ = 2, Z$^2$-ZNCC
performs the worst because of its low data-reuse rate. Furthermore, the larger
the side length, the worse the performance, because step 7 in
Fig.\ref{fig:Z$^2$-ZNCC} does not work effectively.
For $V_Z$ = 5 or 6, the performance is still poor when {\itshape Regcount} = 32
(because of the limited number of registers) or when $H_Z$ = 128 (because of too
much use of registers).
Additionally, for the case shown in Fig.\ref{fig:ZNCC_evaluation} (c), the
larger the $V_Z$ value, the worse the performance of Z$^2$-ZNCC, with the
exception of $V_Z$ = 2.
Comparing Figs.\ref{fig:ZNCC_evaluation} (f) and (i) with (c), we can see that,
by increasing the {\itshape Regcount} to 48 and 60 respectively, the performance
can be improved accordingly.
However, the difference between (f) and (i) is not large, meaning that
increasing {\itshape Regcount} does not improve performance proportionally.
For other cases, the processing time increases with the side length without any
obvious outliers.
The results show that increasing {\itshape Regcount} improves performance more
than changing the $H_Z$, as shown in Fig.\ref{fig:ZNCC_evaluation} (a), (d), and
(g).
Finally, when $V_Z$ = 4, Z$^2$-ZNCC always performs consistently. This means that
it achieves a good balance between hardware resources and calculations according
to our evaluations.
In addition, we also compared Z$^2$-ZNCC to other methods under various maximum
disparity values.
Figure.\ref{fig:ZNCC_compare} shows the results of five methods: two Z$^2$-ZNCC
methods, two original progressive-scan methods, and RTNCC~\cite{r12}. $D$
represents the maximum disparity value and is set to 90, 96 and 128.
Based on the summation-speed comparison, we use Method 1 when the window sizes
are smaller than 9x9 and Method 2 when the window sizes are larger than 7x7.
In the Z$^2$-ZNCC methods, $V_Z$ = 4, $H_Z$ = 32 and {\itshape Regcount} = 60.
The proposed Z$^2$-ZNCC methods work faster than other methods where the
disparity is 96 and 128.
When the side length is 3 and $D$ = 128, the processing time is 18.05 ms for
Original and 11.31 ms for Z$^2$-ZNCC: a 38\% increase in speed.
When the side length is changed to 15, the processing time is 72.09 ms for
Original and 34.35 ms for Z$^2$-ZNCC, meaning that the processing time is
reduced by more than half.
This is mainly because the Z$^2$-ZNCC methods can reuse data in the vertical
direction but the original methods cannot.
Furthermore, compared with the latest RTNCC (its $D$ is only 90), our method for
$D$ = 96 requires only 16.04 ms, which is 26\% faster despite its computational
complexity.
\begin{figure*}
\centering
\includegraphics[width=6.8in]{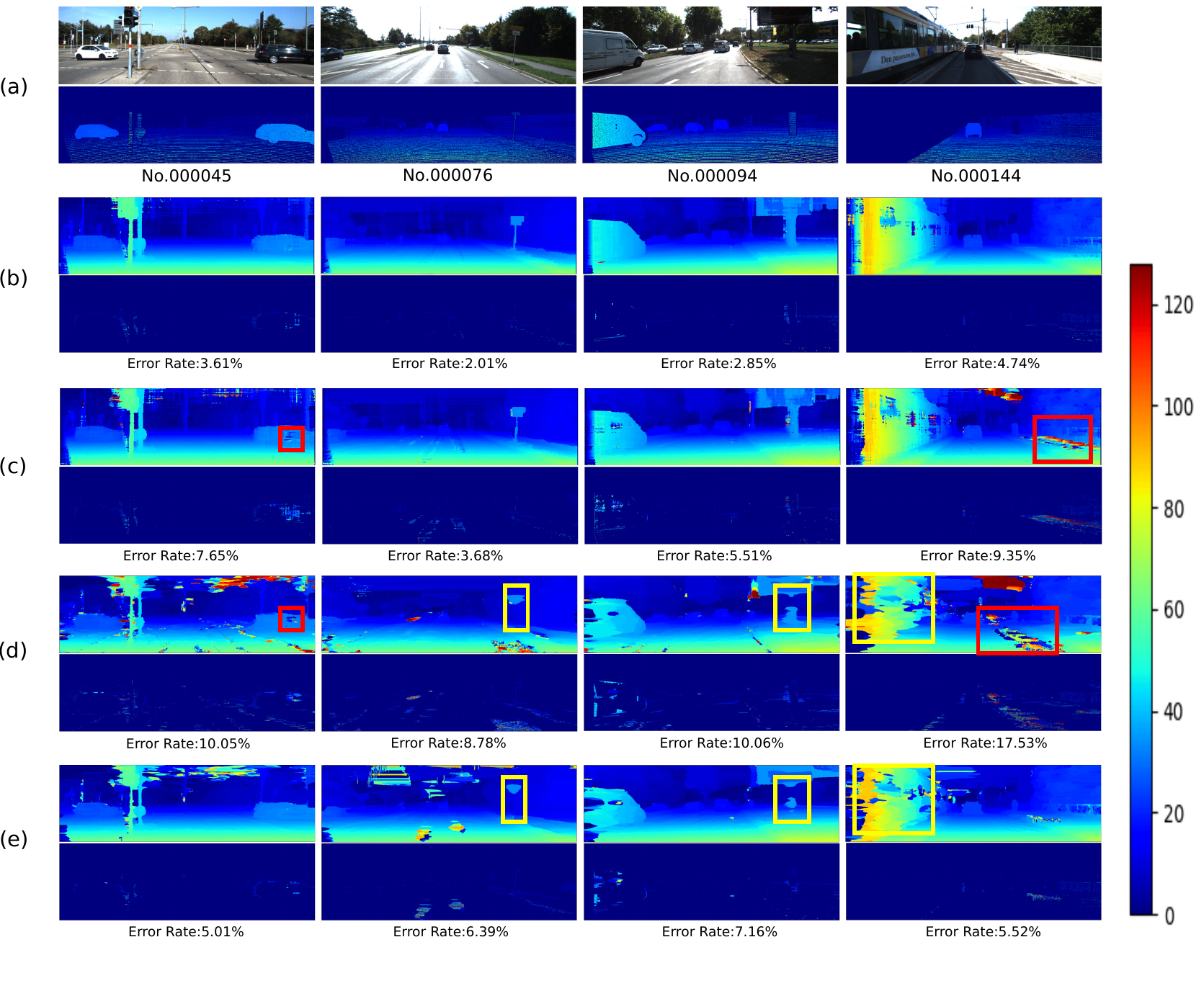}
\vspace*{-0.5cm}
    \caption{Matching Result(KITTI 2015): (a) Left image \& ground truth; (b) S1:
      Z$^2$-ZNCC+SGM; (c) S2: Census+SGM; (d) S3: Census+FastDT;  (e) S4:
      Z$^2$-ZNCC+FastDT}
  \label{fig:DTKITTIresult}
\end{figure*}
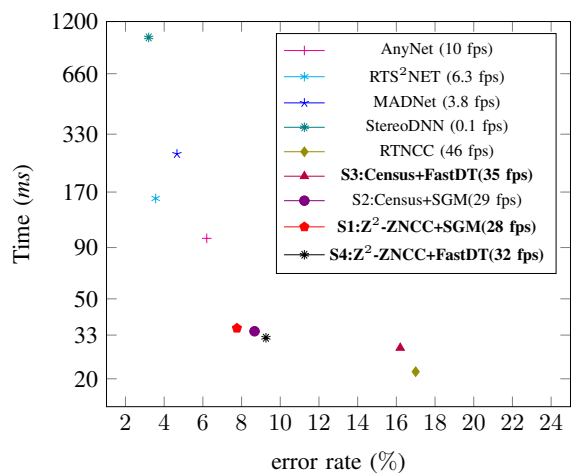
\begin{figure}
\center
\begin{tikzpicture}[scale=0.9]
\begin{axis}
[ xlabel= error rate ($\%$), 
ylabel= Time ({\itshape ms}),
legend style={font=\scriptsize},
legend entries ={AnyNet (10 fps), RTS$^2$NET (6.3 fps), MADNet (3.8 fps), StereoDNN (0.1 fps), RTNCC (46 fps), \bf{S3:Census+FastDT}(35 fps), S2:Census+SGM(29 fps), \bf{S1:Z$^2$-ZNCC+SGM}(28 fps), \bf{S4:Z$^2$-ZNCC+FastDT}(32 fps)},
legend pos = north east,
ymax = 1200,
xmax = 25,
ymode = log,
log basis y={2},
xtick={0,2,...,25},
yticklabels={20,33,50,90,170,330,660,1200},
ytick={20,33,50,90,170,330,660,1200}
] 
\addplot[color=magenta,mark=+] coordinates 
{ 
  (	6.2	,	100	)
};

\addplot[color=cyan,mark=asterisk] coordinates 
{ 
  (	3.56	,	158	)
};

\addplot[color=blue,mark=star] coordinates 
{ 
  (	4.66	,	263.15	)
};

\addplot[color=teal,mark=10-pointed star] coordinates 
{ 
  (	3.2	,	1000	)
};    

\addplot[color=olive,mark=diamond*] coordinates 
{ 
  (	17	,	21.7	)
};

\addplot[color=purple,mark=triangle*] coordinates 
{ 
  (	16.2	,	28.5	)
};

\addplot[color=violet,mark=oplus*] coordinates 
{ 
  (	8.67	,	34.48	)
};

\addplot[color=red,mark=pentagon*] coordinates 
{ 
  (	7.76	,	35.71	)
}; 

\addplot[color=black,mark=10-pointed star] coordinates 
{ 
  (	9.26	,	32	)
};   
\end{axis}
\end{tikzpicture}
\caption{Comparison with other systems}
\label{fig:comparison3}
\end{figure}
\begin{table}
\caption{Comparison of Kernel Performance In Z$^2$ZNCC. }
\begin{tabular}{c|cccc}
\hline
  {\bf Method} & {\bf GT(G/s)} & {\bf GE(\%)} & {\bf IPW} & {\bf IPC} \\
\hline
   Original ZNCC & 0.92 &67.09&1.06e+05&3.05\\
\hline
   Z$^2$ZNCC &1.16&76.95&3.06e+05&3.15\\
\hline
\end{tabular}
\begin{tablenotes}
      \small
      \item GT: gld\_throughput GE: gld\_efficiency IPW: inst\_per\_warp
        IPC:inst\_per\_cycle
\end{tablenotes}
\label{tbl:zncc_bandwidth}
\end{table} 
Table.\ref{tbl:zncc_bandwidth} shows the comparison between
  Z$^2$ZNCC and the original ZNCC in terms of kernel performance. We use {\em
    gld\_throughput} and {\em gld\_efficiency} to evaluate the performance of
  our kernel in memory access, and use {\em inst\_per\_warp} and {\em
    inst\_per\_cycle} for the computational efficiency. In particular, the
  number of our IPW is roughly three times the original, which shows that our
  method can effectively save on-chip resources to maintain a high degree of
  thread parallelism.

\subsection{Evaluation of FastDT}
\label{sec:5_3}
We evaluated FastDT in combination with Z$^2$-ZNCC using the {\itshape training
  set} of the KITTI 2015~\cite{r6} benchmark because the accuracy must be
evaluated precisely.
200 pairs of images are included in the {\itshape training set}, and their sizes
are close to 1,250x375 with a maximum disparity value of 128.
In our evaluation, {\itshape Regcount} = 48, {\itshape T} = 21 and the side
length of ZNCC is set to 3. The two parameters in \eqref{equ:weight} are
$\sigma_s$ = 5 and $\sigma_r$ = 52. $d_{arb}$ is set to 0 to effectively reduce
the range of cost values because $d_{min}$ is rarely equal to zero in stereo
matching.
\begin{table}
\caption{Comparison of DT in KITTI 2015. }
\begin{tabular}{p{0.12\textwidth}p{0.045\textwidth}p{0.045\textwidth}p{0.045\textwidth}p{0.045\textwidth}p{0.045\textwidth}}
\hline
  {\bf Direction} & {\bf Float} & {\bf Int32} & {\bf Int16} & {\bf Int8 (Ours)} & {\bf Int16-NP} \\ 
\hline
L2R$^{*}$(ms) & 13.10 & 13.76 & 13.63 & 13.68 & 13.59 \\
R2L(ms) & 15.41 & 15.32 & 8.72 & 5.52& 8.56  \\
U2D(ms) & 15.12 & 15.05 & 8.74 & 7.22& 8.65  \\
D2U$^{*}$(ms)  & 7.06 & 7.03 & 6.29  & 6.27& 6.34 \\
Frame Rate(fps) & 20 & 20 & 26 & 32 & 27 \\ 
Error Rate(\%) & 7.36 & 7.49 & 7.57 & 7.63 & 8.14  \\ 
\hline
\end{tabular}
\begin{tablenotes}
      \small
      \item L2R$^*$:Z$^2$-ZNCC \& left to right. R2L:
  right to left.  U2D: up to down. D2U$^*$: down to up \& WTA. NP: non-normalization.
\end{tablenotes}
\label{tbl:DT_comparison}
\end{table} 

Table \ref{tbl:DT_comparison} shows the processing time for each step of DT
under four different methods, together with their accuracy.
The four methods--{\itshape Float}, {\itshape Int32}, {\itshape Int16}, and
{\itshape Int8}--use floating-point, integer, short integer, and character
respectively during cost propagation.
Because Z$^2$-ZNCC calculates the cost value for each pixel serially, it is
combined with the cost propagation from left to right (L2R).
WTA is also combined with the cost propagation from down to up (D2U) to
calculate the disparity map directly without transferring cost values back to
the global memory.
According to our observation, the cost values aggregated during the L2R step are
not usually large. Therefore, in our implementation, cost-value normalization is
performed only in the U2D steps of {\itshape Float}, {\itshape Int32}, and
{\itshape Int16}, and in the R2L step of {\itshape Int8}.
The encoding is performed at the end of the R2L step and
the decoding is performed at the beginning of the U2D step.
Additionally, to verify the necessity of cost-value normalization, we added the
evaluation of {\itshape Int16-NP}, in which such normalization is not performed.
In the L2R step, the processing time is roughly the same regardless of the data
type because the transmission latency is hidden by the calculation time of
Z$^2$-ZNCC.
On the other hand, the processing time of {\itshape Int16} is only 8.72 ms in
the R2L step and 8.74 ms in the U2D step, which are roughly half of the values
for {\itshape Float} and {\itshape Int32}.
For {\itshape Int16-NP}, since it is roughly the same as {\itshape Int16} in
terms of calculation and data transmission, there is no obvious difference in
processing time.
Our {\itshape Int8} shows the fastest processing speed of all methods, even
though data encoding \& decoding is performed.
This is because our method transfers data in 8-bit, which greatly reduces the
latency of data transfer.
Table.\ref{tbl:DT_bandwidth} shows the comparison between FastDT
  and the original DT in terms of global memory accessing. As the results of
  {\em gld\_throughput} and {\em gld\_efficiency} shown, our kernel
  significantly improves the efficiency of global memory access in both {\em
    R2L} and {\em U2D} directions. This is mainly benefits from our encoder
  compression method, which can improve transmission efficiency by bundling
  data.
In our implementation, {\itshape Int8} achieved the requirement for real-time
processing with 32 fps, which is 60\% faster than the 20 fps attained by
{\itshape Float}.

\begin{table}
\caption{Comparison of Global Memory Efficiency In DT. }
\begin{tabular}{c|c|cccc}
\hline
  \multicolumn{2}{c|}{\bf Method} & {\bf L2R} & {\bf R2L} & {\bf U2D} & {\bf D2U} \\
\hline
\multirow{2}*{Original DT} & GT(G/s) &0.42&0.56&5.18&38.7 \\
&GE(\%)&50.3&20.29&59.39&94.16\\
\hline
\multirow{2}*{FastDT} & GT(G/s) &0.41&11&17.08&22.6 \\
&GE(\%)&50.13&99.2&88.9&88.9\\
\hline
\end{tabular}
\begin{tablenotes}
      \small
      \item GT: gld\_throughput GE: gld\_efficiency.
\end{tablenotes}
\label{tbl:DT_bandwidth}
\end{table}

\begin{table*}
  \caption{Comparison of matching accuracies in KITTI 2015}
  \vspace*{-0.2cm}
\begin{center}
\begin{tabular}{p{0.12\textwidth}|p{0.045\textwidth}p{0.045\textwidth}p{0.05\textwidth}p{0.045\textwidth}|p{0.045\textwidth}p{0.045\textwidth}p{0.05\textwidth}p{0.045\textwidth}|p{0.045\textwidth}p{0.045\textwidth}p{0.05\textwidth}p{0.045\textwidth}}
  \hline
  \textbf{Error Rate (\%)} &&&\textbf{D1-bg}&&&&\textbf{D1-fg}&&&&\textbf{D1-all}\\
  \hline
  &\textbf{S1}&\textbf{S2}&\textbf{S3}&\textbf{S4}&\textbf{S1}&\textbf{S2}&\textbf{S3}&\textbf{S4}&\textbf{S1}&\textbf{S2}&\textbf{S3}&\textbf{S4}\\
  \hline
  All / All & 6.60 & 6.98 & 15.12 & 7.77 & 13.58 & 17.04 & 23.56 & 16.71 & 7.76 & 8.66 & 16.52 & 9.26  \\
  All / Est & 6.54 & 6.82 & 12.56 & 7.61 & 13.54 & 16.92 & 22.78 & 16.52 & 7.71 & 8.51 & 14.24 & 9.10  \\
  Noc / All & 5.22 & 5.48 & 14.33 & 6.92 & 11.38 & 14.83 & 21.88 & 15.01 & 6.24 & 7.03 & 14.48 & 8.26  \\
  Noc / Est & 5.20 & 5.44 & 11.73 & 6.87 & 11.38 & 14.82 & 21.38 & 14.99 & 6.22 & 6.99 & 13.38 & 8.21  \\
  \hline
\end{tabular}
\begin{tablenotes}
      \small
      \item S1: Z$^2$-ZNCC+SGM. S2: Census+SGM. S3: Census+FastDT. S4:
        Z$^2$-ZNCC+FastDT.
\end{tablenotes}
\label{Table:accuracy}
\end{center}
\vspace*{-0.3cm}
\end{table*} 
In the accuracy evaluation, {\itshape Float} has the lowest error rate of
8.16\%. As the data width decreases, the error rate gradually increases.
Compared with the error rate of 8.37\% for {\itshape Int16}, that for {\itshape
  Int16-NP} is higher (8.81\%) because of the overflow of the cost values.
This shows that our cost normalization with nearly zero-mean works very
effectively.
{\itshape Int8} has an error rate of 8.41\%, only a 0.25\% loss compared to
{\itshape Float}. This means that our method can effectively increase the
processing speed even as it maintains a high accuracy in stereo matching.
\subsection{Evaluation of Stereo Matching}
\label{sec:5_4}
Finally, to further clarify the effectiveness of our Z$^2$-ZNCC and FastDT for
stereo matching, we also combined them with the state-of-the-art algorithms SGM
and census, respectively.
Then, we compared their accuracies and processing speeds on a Jetson Tx2 GPU
using the KITTI 2015 benchmark.
SGM and census were chosen because the GPU system in~\cite{r7} uses them and has
a good performance (8.66\% error rate, 29 fps) under the same conditions.
Our implementation of the SGM is almost the same as~\cite{r7}. The difference is
that to clarify the role of Z$^2$-ZNCC, we do not use the {\itshape stream}
function.
The two parameters in SGM--$P1$ and $P2$--are set to 18 and 185, respectively,
because the results of Z$^2$-ZNCC are multiplied by the coefficient $T$.

Figure.\ref{fig:DTKITTIresult} shows the comparisons of three systems based
on use our proposed methods, and one other system~\cite{r7}.
The four systems are Z$^2$-ZNCC+SGM, Census+SGM~\cite{r7}, Census+FastDT and
Z$^2$-ZNCC+FastDT; they are expressed as {\itshape S1}, {\itshape S2}, {\itshape
  S3}, and {\itshape S4}, respectively.
Four pairs of images are selected from the {\itshape training set} to clearly
show the difference in the results of these systems.
Figure.\ref{fig:DTKITTIresult} (a) shows the reference images and their
ground truths.
Figures.\ref{fig:DTKITTIresult} (b) to \ref{fig:DTKITTIresult} (e) show the results
of {\itshape S1} to {\itshape S4}, respectively.
The top half of each figure shows disparity map and the bottom half the
corresponding error image compared with the ground truths.
The color bar denotes the disparity range of [0,128), with blue representing the
  farthest objects and red the closest.
As for the four sets of results No.000045, No.000076, No.000094, and No.000144,
{\itshape S1} in Fig.\ref{fig:DTKITTIresult} (b) shows a clear advantage in
terms of accuracy. Its error rates of 3.61\%, 2.01\%, 2.85\%, and 4.75\% are
obviously lower than those of other systems.
Compared with {\itshape S2} in Fig.\ref{fig:DTKITTIresult} (c), {\itshape S1}
works better in photometric distortions and weak-pattern areas, as shown by the
red boxes in Figs.\ref{fig:DTKITTIresult} (c) and (d).
This means that ZNCC has stronger robustness than census.
{\itshape S3} shows a disadvantage in accuracy. Its error rates are
the highest among the four systems.
This is because, in addition to the less accurate matching by census, DT also
easily causes a fattening effect, making some details disappear, as shown in the
yellow boxes in Figs.\ref{fig:DTKITTIresult} (d) and (e).
By replacing census with ZNCC, the accuracy of {\itshape S4} has been greatly
improved, as shown in Fig.\ref{fig:DTKITTIresult} (e).
This means that as long as a high precision is achieved in the cost matching
stage, DT will not induce an excessive loss in accuracy.
Table \ref{Table:accuracy} compares the matching accuracies of the four systems
using the {\itshape testing set} of KITTI 2015 benchmark. The result is
consistent with the above, and the order of accuracy is {\itshape
  S1}$>${\itshape S2}$>${\itshape S4}$>${\itshape S3}. The use of ZNCC improves
the accuracies of the census-based systems by 0.9\% ({\itshape S1} vs. {\itshape
  S2}) and 7.26\% ({\itshape S4} vs. {\itshape S3}).

Figure.\ref{fig:comparison3} shows a comparison with other systems in terms of
processing speed and accuracy.
The $x$-axis represents the error rate and the $y$-axis represents the time
required for processing in {\itshape ms}; thus, the closer the evaluation
results are to the origin, the higher the accuracy and the faster the processing
speed.
All of the CNN-based systems (AnyNet~\cite{r3}, StereoDNN~\cite{r4},
MADNet~\cite{r5}, and RTS$^2$Net~\cite{r19}) achieved high accuracy, but had no
speeds exceeding 10 fps. RTNCC~\cite{r12} and our {\itshape S3} (Census+FastDT)
algorithm achieved real-time processing speeds of 46 fps and 35 fps,
respectively. However, their error rates are larger than 16\%, which also limits
their usability. {\itshape S2} (Census+SGM)~\cite{r7} and our {\itshape S1}
(Z$^2$-ZNCC+SGM) have almost the same processing speed, but our accuracy is
0.5\% lower. This shows that our method plays a significant role in stereo
matching.
As mentioned above, our {\itshape S4} (Z$^2$-ZNCC+FastDT) improves the accuracy
of {\itshape S3} from 16.52\% to 9.26\%. Compared with the systems based on SGM
({\itshape S1} and {\itshape S2}), {\itshape S4} still has a 1\% lower accuracy,
but its processing speed is 17\% faster, allowing it to truly achieve real-time
processing.
\section{Conclusion}
\label{sec:6}
In this paper, we proposed an acceleration method for the {\itshape zero-means
  normalized cross correlation (ZNCC)} template-matching algorithm for stereo
vision on an embedded GPU.
This method helps us reuse intermediate calculation results efficiently without
frequently transferring them among the hierarchy of memories, leading to a
higher processing speed.
It also helps to improve the matching accuracy because of its
stronger robustness.
We evaluated our systems based on this method by using the KITTI 2015 benchmark
and showed their efficiency.
Since the proposed method does not rely upon the GPU experimental architecture,
it can be used on different GPUs; furthermore, it is not limited to GPUs and can
be used on other hardware platforms like FPGAs.

To further improve out systems' performance, we are planning to combine it with
other cost-aggregation algorithms. This will be done in future work.

\end{document}